\documentclass[
]{ceurart}

\usepackage{listings}
\usepackage{xcolor}
\usepackage{hyperref}
\usepackage[inline]{enumitem}

\usepackage{graphicx}
\usepackage{multirow}

\usepackage{subcaption}

\usepackage{verbatim}
\usepackage{newunicodechar}

\usepackage{wrapfig}

\usepackage{mdframed}
\usepackage{stfloats}

\usepackage{array}
\newcolumntype{C}[1]{>{\centering\arraybackslash}p{#1}}

\usepackage{tcolorbox}
\newcounter{promptno}[section]
\newlength\mystoreparindent
\newenvironment{prompt}[1][]
{
  \setlength{\mystoreparindent}{\the\parindent}
  \setlength{\parindent}{0pt}
  \refstepcounter{promptno}
  \par\medskip
  \noindent
  \begin{tcolorbox}[left=1pt,right=1pt]
  \textsc{{template \small\thesubsection.\thepromptno}}\\
  \small
  \tt
}{
  \end{tcolorbox}
  \setlength{\parindent}{\mystoreparindent}
  \medskip
}

\begin{document}

\copyrightyear{2022}
\copyrightclause{Copyright for this paper by its authors.
  Use permitted under Creative Commons License Attribution 4.0
  International (CC BY 4.0).}

\conference{1st Workshop on Interactive Task Learning in Human-Robot co-construction (ITL4HRI) | IEEE RO-MAN 2025: International Conference on Robot and Human Interactive Communication 2025, August 25–29, 2025, Eindhoven, Netherlands%ITL4HRI Workshop | IEEE RO-MAN 2025: International Conference on Robot and Human Interactive Communication 2025,
  %August 25--29, 2025, Eindhoven, Netherlands
  }

\title{Towards No-Code Programming of Cobots: Experiments with Code Synthesis by Large Code Models for Conversational Programming}

\author[1]{Chalamalasetti Kranti}[%
email=kranti.chalamalasetti@uni-potsdam.de,
]
\cormark[1]
% \address[1]{Computational Linguistics, Department of Linguistics, University of Potsdam, Germany}

\author[1]{Sherzod Hakimov}[%
email=sherzod.hakimov@uni-potsdam.de,
]
% \address[1]{Computational Linguistics, Department of Linguistics, University of Potsdam, Germany}

\author[1, 2]{David Schlangen}[%
email=david.schlangen@uni-potsdam.de,
]
\address[1]{Computational Linguistics, Department of Linguistics, University of Potsdam, Germany}
\address[2]{German Research Center for Artificial Intelligence (DFKI), Berlin, Germany}

\cortext[1]{Corresponding author.}

\begin{abstract}
While there has been a lot of research recently on robots in household environments, at the present time, most robots in existence can be 
found on shop floors, and most interactions between humans and robots happen there. 
``Collaborative robots'' (cobots) designed to work alongside humans on assembly lines traditionally require expert programming, limiting ability to make changes, or manual guidance, limiting expressivity of the resulting programs. To address these limitations, we explore using Large Language Models (LLMs), and in particular, their abilities of doing in-context learning, for conversational code generation. As a first step, we define SARTCo, the `` Spatial Arrangement and Reconstruction Tasks for Cobots '', a 2.5D structure building task designed to lay the foundation for simulating industry assembly scenarios. In this task, a `programmer' instructs a cobot, using natural language, on how a certain assembly is to be built; that is, the programmer induces a program, through natural language. We create a dataset that pairs target structures with various example instructions (human-authored, template-based, and model-generated) and example code. With this, we systematically evaluate the capabilities of state-of-the-art LLMs for synthesising this kind of code, given in-context examples. Evaluating in a simulated environment, we find that LLMs are capable of generating accurate `first order code' (instruction sequences), but have problems producing `higher-order code' (abstractions such as functions, or use of loops).
\end{abstract}


\maketitle

\section{Introduction}
	
Collaborative robots (cobots), designed to work safely alongside humans, have traditionally required expert programming~\citep{DBLP:journals/ras/ZaatariMLU19, DBLP:conf/ro-man/GiannopoulouBM21}, hindering wider accessibility for novice workers. To bridge this gap, conversational programming is emerging as a possible solution, demanding systems that can parse human input and craft corresponding programs~\citep{DBLP:journals/corr/abs-2003-01318} interactively. While historically methods have relied on domain-specific model training and learning by demonstration, these techniques often fall short due to their extensive data needs and inability to handle complex instructions~\citep{DBLP:journals/ijhr/BauerWB08, DBLP:journals/rcim/MukherjeeGCN22, DBLP:journals/trob/RozoCCJT16,berg2020review}. This necessitates exploring efficient alternative approaches for cobot programming, such as program synthesis.

\begin{figure}
\centering
  \includegraphics[width=\textwidth]{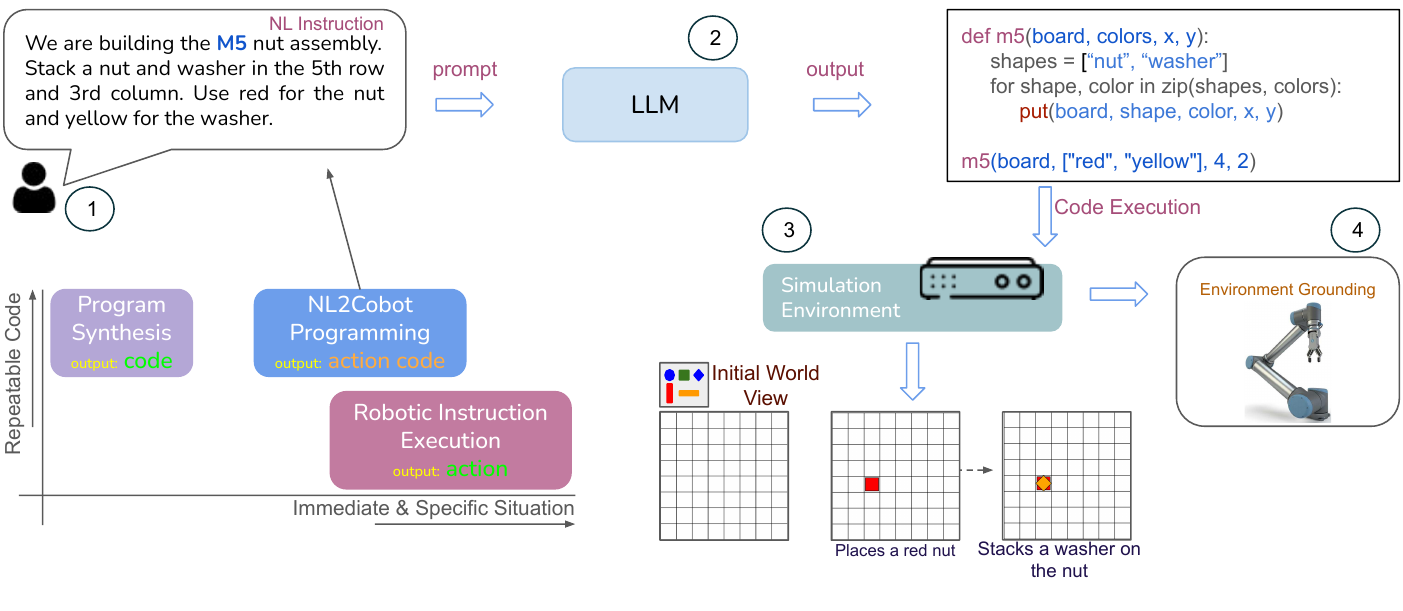}
  \caption {Overview of NL2Cobot Programming, demonstrating the abstraction of user instructions into a high-level executable function and its execution in a simulated environment.}
  \label {fig:overview}
\end{figure}

Program synthesis from natural language instructions (NL2Code) is an active research area in Natural Language Processing, leveraging LLMs for tasks like code completion, debugging, and generating programs from natural language descriptions~\citep{DBLP:journals/corr/abs-2107-03374, DBLP:conf/iclr/NijkampPHTWZSX23, wei2023magicoder, DBLP:journals/corr/abs-2401-14196}. In robotics, LLMs have been increasingly used for task planning~\citep{DBLP:conf/corl/IchterBCFHHHIIJ22, DBLP:conf/icml/HuangAPM22, DBLP:conf/iclr/ZengAICWWTPRSLV23, DBLP:journals/corr/abs-2303-00855}, grounding~\citep{DBLP:journals/corr/abs-2210-03094, DBLP:conf/nips/Huang0WHSML0MPL23, DBLP:conf/icml/DriessXSLCIWTVY23}, and instruction following, focusing primarily on immediate, specific actions~\citep{DBLP:conf/icra/LiangHXXHIFZ23, DBLP:journals/arobots/SinghBMGXTFTG23, 10473591, DBLP:journals/corr/abs-2310-10645, DBLP:conf/corl/LiuYILSTS23} for ``here and now" scenarios. However, replicating these instructions in new environments require generating a new set of actions each time, which can be inefficient and cumbersome. Instead, abstracting robot actions as higher-order programs makes them adaptable, repeatable, and generalizable to novel environments, significantly improving flexibility and efficiency, which is essential for industrial contexts. Our research (see Figure~\ref{fig:overview}) sits between traditional program synthesis (generating code for programmers who understand it) and robot instruction following (generating robot actions, not repeatable programs).

The proposed work focuses on the step from natural language to programs for building assemblies in a construction setting. As such, it is related to that of \citet{paetzel2022conversational}; however, we abstract away questions of the physical execution of the programs on a robot and instead simulate the setting and the effects of pick and place operations.

Our key contributions are: (a) a component assembly dataset with different categories of objects and styles of instructions (Section~\ref{sec:taskenv}); (b)  an evaluation methodology to assess program synthesis capabilities of LLMs (Section~\ref{sec:expsetup}); (c) an experimental setup and comprehensive analysis of differences in program synthesis behavior between commercial and open-access code generation LLMs (also Section~\ref{sec:expsetup})

\section{Related Work}

\paragraph{Program Synthesis:} There exist several natural-language-to-code (NL2Code) datasets~\citep{DBLP:journals/corr/abs-1709-00103, DBLP:conf/nips/LuGRHSBCDJTLZSZ21, DBLP:journals/corr/abs-2108-07732, DBLP:journals/corr/abs-2211-15533} for training machine learning models~\citep{DBLP:journals/corr/abs-2107-03374, DBLP:conf/iclr/NijkampPHTWZSX23, DBLP:journals/corr/abs-2307-09288, DBLP:journals/corr/abs-2308-12950, wei2023magicoder, DBLP:journals/corr/abs-2401-14196}. Numerous studies have utilized these datasets to explore the capabilities of LLMs, as surveyed by~\citep{DBLP:journals/corr/abs-2308-10620, DBLP:journals/entropy/WongGHHT23, DBLP:conf/acl/ZanCZLWGWL23}. While these studies demonstrate the ability of LLMs to generate and complete code, they predominantly target users familiar with programming concepts and terminologies. In contrast, our work focuses on code generation in an industrial context, where the end-users are novice workers without programming backgrounds. This raises challenges in dealing with ambiguous instructions, domain-specific knowledge that LLMs might not be aware of, and insufficient context, making it a necessary area for exploration.

\paragraph{Building Tasks:} In the virtual block world of Minecraft, datasets have emerged from structure-building tasks~\citep{DBLP:conf/acl/JayannavarNH20, DBLP:conf/emnlp/BaraCC21, DBLP:journals/corr/abs-2211-00688}, linking instructions with actions in a 3D grid. These datasets resemble industry-assembly tasks involving detailed instructions and actions. However, these tasks' high complexity, lack of spatial constraints and intricate instructions challenge even the best current LLMs~\citep{DBLP:journals/corr/abs-2402-08392}, limiting their applicability to such collaboarative 3D building tasks. Furthermore, these datasets lack tasks involving repeating target structures, which is common in cobot applications. On the other hand, LLMs have shown promising performance in abstract spatial tasks~\citep{DBLP:conf/corl/MirchandaniXFID23, DBLP:journals/corr/abs-2305-18354} in 2D controlled environment, highlighting their abstraction, reasoning and execution capabilities. This discrepancy inspired us to propose a 2D building task that is tailored for controlled industrial environments and includes repetitive tasks and giving us an opportunity to measure different aspects of LLMs abilities in synthesizing code.

\paragraph{Instruction-Following Tasks:} A favored domain in this space is human-agent interaction for household tasks~\citep{DBLP:conf/cvpr/ShridharTGBHMZF20, DBLP:conf/aaai/PadmakumarTSLNG22, DBLP:journals/corr/abs-2403-02274}, where the focus is on a single execution from a given starting point of command. This is typically framed as generating a sequence of actions (and hence a program which is not intended to be repeated). Similar to our setting, \textit{HEXAGONS} dataset~\cite{DBLP:journals/tacl/LachmyPMT22} aims to translate natural language instructions into programs potentially using higher-level constructs such as loops. However, this dataset focuses more on simulating drawing rather than constructions; as we will see, our dataset puts more weight on introducing assemblies that, in turn, can be built out of smaller assemblies.

\paragraph{Language to Robot Programs:} AutoMisty~\citep{DBLP:journals/corr/abs-2503-06791} and \citet{DBLP:conf/icra/MacalusoCC24} demonstrate the use of LLMs for code generation in robotics tasks, with a focus on decomposing high-level instructions into subtasks prior to code synthesis. RoboCodeX~\citep{DBLP:conf/icml/0001CZCYGCLHTSY24} explores the use of multimodal LLMs for robotic manipulation planning by translating natural language instructions and visual observations into structured manipulation sequences. Their focus lies in inferring object-centric manipulation actions. ROBO-INSTRUCT~\citep{DBLP:journals/corr/abs-2405-20179} is a framework focused on synthetic robot program generation useful for fine-tuning open-weight LLMs. Unlike these works, our setup isolates the code generation problem by providing natural language instructions of varying complexity, and evaluating the model’s ability to generate correct executable programs without requiring additional planning or interpretation.

\section{Simulating Industry-Style Assembly with 2.5D Building Tasks}
\label{sec:taskenv}

\subsection{Task Overview}
\label{subsec:taskoverview}
\paragraph{Blocks World on Grid} SARTCo~\footnote{Pronounced as SART-ko} is inspired by traditional grid-based Blocks World problems but introduces two important differences. First, instead of using a single type of object (i.e., a `block'), SARTCo features a variety of component types. Second, while standard Blocks World elements typically occupy a single cell, lack orientation, and can be placed or stacked freely, components in SARTCo may span multiple cells, have specific orientations, and must adhere to defined stacking rules.

\paragraph{Structure Building Task} Our proposed 2.5D~\footnote{A 2D grid where components can also be stacked vertically, without the complexity of full 3D simulation.} building task is inspired by HEXAGONS \cite{DBLP:journals/tacl/LachmyPMT22} and MINECRAFT Collaborative Building \citep{DBLP:conf/acl/JayannavarNH20}. We follow the theme of these datasets and define the task as an interaction between an instruction giver (the programmer) and an instruction follower (the cobot). The programmer, which can be a human, a model, or a rule-based generator, is provided with a target board and must instruct the cobot (an instruction-tuned code generation LLM) using natural language to reconstruct the target board while adhering to spatial constraints. The programmer is limited to generating instructions, and the cobot is limited to generating code. Since the cobot is designed to focus solely on code generation, it does not engage in clarification-seeking during the interaction. We plan to explore support for clarification behavior in future iterations of the dataset~\footnote{Dataset is available at: \url{https://github.com/clp-research/cobotprogsynthesis-coderep}}.

Unlike prior datasets where the programmer can observe and respond to the cobot’s intermediate actions, our dataset is intentionally non-interactive. The programmer does not receive execution feedback while providing instructions. To support different instruction styles, the dataset includes both multi-turn and single-turn variants. In the multi-turn setting, instructions are provided incrementally across turns. In the single-turn setting, all instructions are provided at once. In both cases, the execution of the generated code occurs only at the end. Hence the programmer cannot view any output during the interaction. This design deliberately avoids partial execution feedback, preventing the programmer from making real-time adjustments based on the cobot’s progress. We plan to explore interactive variants in future versions, where intermediate visualization and instruction correction will be supported.

This non-interactive structure aligns with a compiler-and-executor paradigm, where the full set of instructions is treated as a complete program to be executed in its entirety. By removing feedback mechanisms, the dataset prevents mid-task corrections. As a result, all instructions must be perspective agnostic. They cannot rely on relative spatial terms such as ``to your left'' or ``in front of you'', which depend on a shared or shifting viewpoint. Instead, spatial references must be grounded in absolute grid coordinates to ensure that the cobot can interpret them unambiguously. The task concludes once the programmer has provided the full instruction sequence.

\begin{figure*}[t]
  \includegraphics[width=\textwidth]{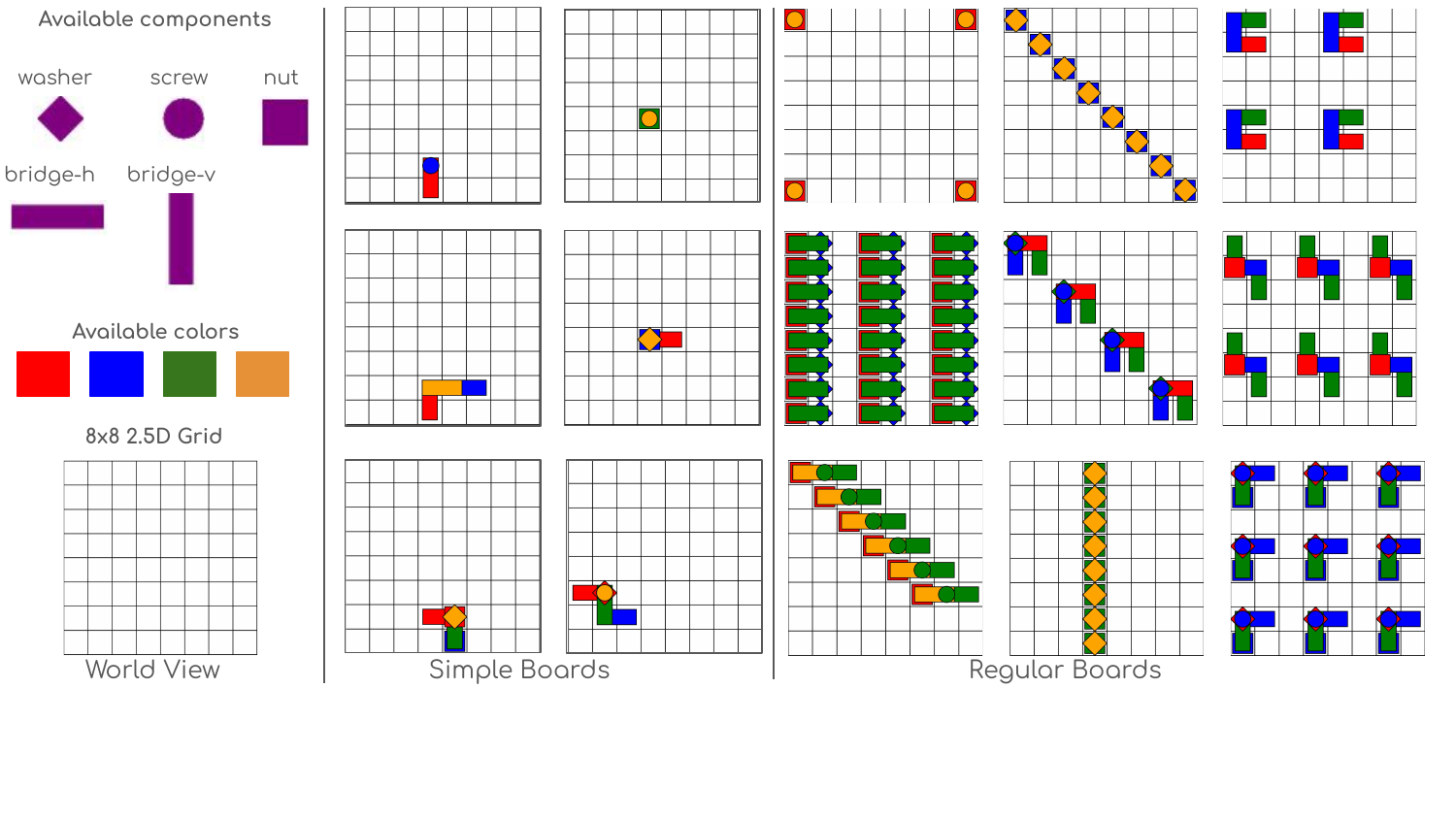}
  \caption {Overview of the environment with available components, and samples of simple and regular boards.}
    \label{fig:boards_overview}
    \vspace{-0.50cm}    
\end{figure*}

\paragraph{Industry-Style Assembly} To mimic industrial assembly scenarios, we model the components as four types: \textit{washers}, \textit{nuts}, \textit{screws}, and \textit{bridges} (horizontal or vertical), each available in one of four colors: \textit{red, green, blue, and yellow}. The cobot is equipped with an unlimited inventory of these component types across all colors. The placement of components adheres to a set of predefined rules that reflect real-world assembly logic; for instance, a screw must always be placed on top, consistent with standard assembly conventions. A full description of the placement constraints is provided in Appendix~\ref{sec:appendix_simul_env}.

We define a specific arrangement of connected components as an object. Two or more components are considered part of the same object if they are connected in a way that would conceptually allow electricity to flow through them. A grid containing one or more such objects is referred to as a board. We categorize boards into two types: simple boards, which contain a single object, and regular boards, which feature repetitions of the same object `\textit{n}' times arranged in varying patterns. Regular boards are further divided into two categories based on the size of the base structure: simple regular boards, where the base structure occupies a single cell, and complex regular boards, where the base structure spans multiple cells. Examples of both board types are shown in Figure~\ref{fig:boards_overview}.

\subsection{Dataset Overview}
\label{subsubsec:boardgen}
\subsubsection{Target Board Generation} 
\begin{figure*}[t]
\centering
  \includegraphics[width=\textwidth]{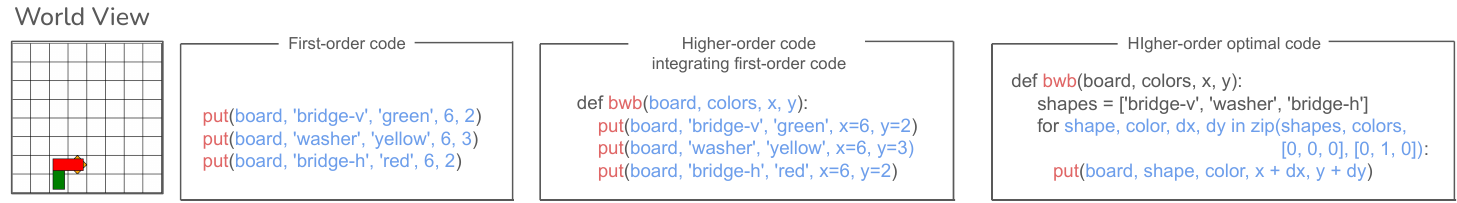}
  \caption {Various gold standard code styles represent the target structure: (a) features first-order code snippet using primitive controls, (b) code includes higher-order functions built from this first-order code, and (c) code featuring optimal higher-order functions.}
    \label{fig:code-cats}
\end{figure*}

To facilitate the automatic creation of objects, we represent the 2.5D grid as an array. This array-based representation allows for control and manipulation of shape placements. By abstracting the arrangement of components on the grid into a Python program, we can generate numerous objects. To control task complexity, we defined a limited number of such python programs (referred to as seeds, and expressed using \textit{Jinja2\footnote{\url{https://palletsprojects.com/p/jinja/}}} templates) on an 8x8 grid and extrapolated them with all the possible combinations of components, colors and locations. Table~\ref{tab:boardseeds} details these seeds and the number of possible objects and boards across the four quadrants of the grid. The code snippet representing each board serves as the gold-standard code during the evaluation. We further expand these programs to generate two additional forms using \textit{Jinja2} templates (see Figure~\ref{fig:code-cats}): (a) \textit{first-order code}, involving a series of \textit{put} commands, and (b) a \textit{higher-order function} that integrates these first-order code sequences.

Although the structures in simple boards may appear straightforward, they require models to accurately identify colors, extract spatial positions, associate them with the correct shapes, and determine the correct order of placement. In contrast, regular boards demand higher-level pattern understanding, such as recognizing diagonal arrangements or alternating rows and columns. This involves not only interpreting the pattern described in the instructions but also generating code that reproduces it correctly. For example, even if the pattern specifies a diagonal layout, the model must reason that the next object cannot simply go in the next row if the base structure spans multiple cells. Collectively, these boards serve as a testbed for evaluating a model's reasoning ability and generalization across different spatial and structural configurations.

\begin{table}
\footnotesize
\caption{Distribution of generated boards across all categories; NS: number of seeds, used to generate the boards, NO: number of distinct objects, NB: number of boards generated using all color and shape combinations}
\label{tab:boardseeds}
\begin{tabular}{|c|c|c|c|c|}
\hline
Board Type & Object Type & NS & NO & NB \\ \hline
Simple & Simple & 18 & 68 & 152,352 \\ \hline
\multirow{2}{*}{Regular} & Simple & 5 & 68 & 45,984 \\ \cline{2-5} 
 & Complex & 10 & 68 & 156,948 \\ \hline
\end{tabular}
\end{table}

\begin{table}
\footnotesize
\caption{Breakdown of board and object types for training, validation, and test splits. Out of all the possible boards available (see Table~\ref{tab:boardseeds}), these samples were randomly selected.}
\label{tab:dataset_stats}
\begin{tabular}{|c|c|ccc|}
\hline
\multirow{2}{*}{Board Type} & \multirow{2}{*}{Object Type} & \multicolumn{3}{c|}{Total Boards} \\ \cline{3-5} 
 &  & \multicolumn{1}{c|}{Train} & \multicolumn{1}{c|}{Val} & Test \\ \hline
Simple & Simple & \multicolumn{1}{c|}{1072} & \multicolumn{1}{c|}{130} & 130 \\ \hline
\multirow{2}{*}{Regular} & Simple & \multicolumn{1}{c|}{1168} & \multicolumn{1}{c|}{130} & 130 \\ \cline{2-5} 
 & Complex & \multicolumn{1}{c|}{2944} & \multicolumn{1}{c|}{130} & 130 \\ \hline
\end{tabular}
\end{table}

\paragraph{Dataset splits} Out of all the possible boards available (see Table~\ref{tab:boardseeds}), we randomly selected the samples (see Table~\ref{tab:dataset_stats}) for evaluation purposes. The selected samples are divided into training, validation, and test splits. All data splits contain at least one board for each possible component combination. All training split samples feature objects in the 8x8 grid’s top-left quadrant, validation samples in the top-right, and test samples in either bottom quadrant, ensuring coverage of all grid areas.

\subsubsection{Instruction Generation}
\label{subsubsec:instructiongen}

\begin{figure*}[t]
\centering
  \includegraphics[width=\textwidth]{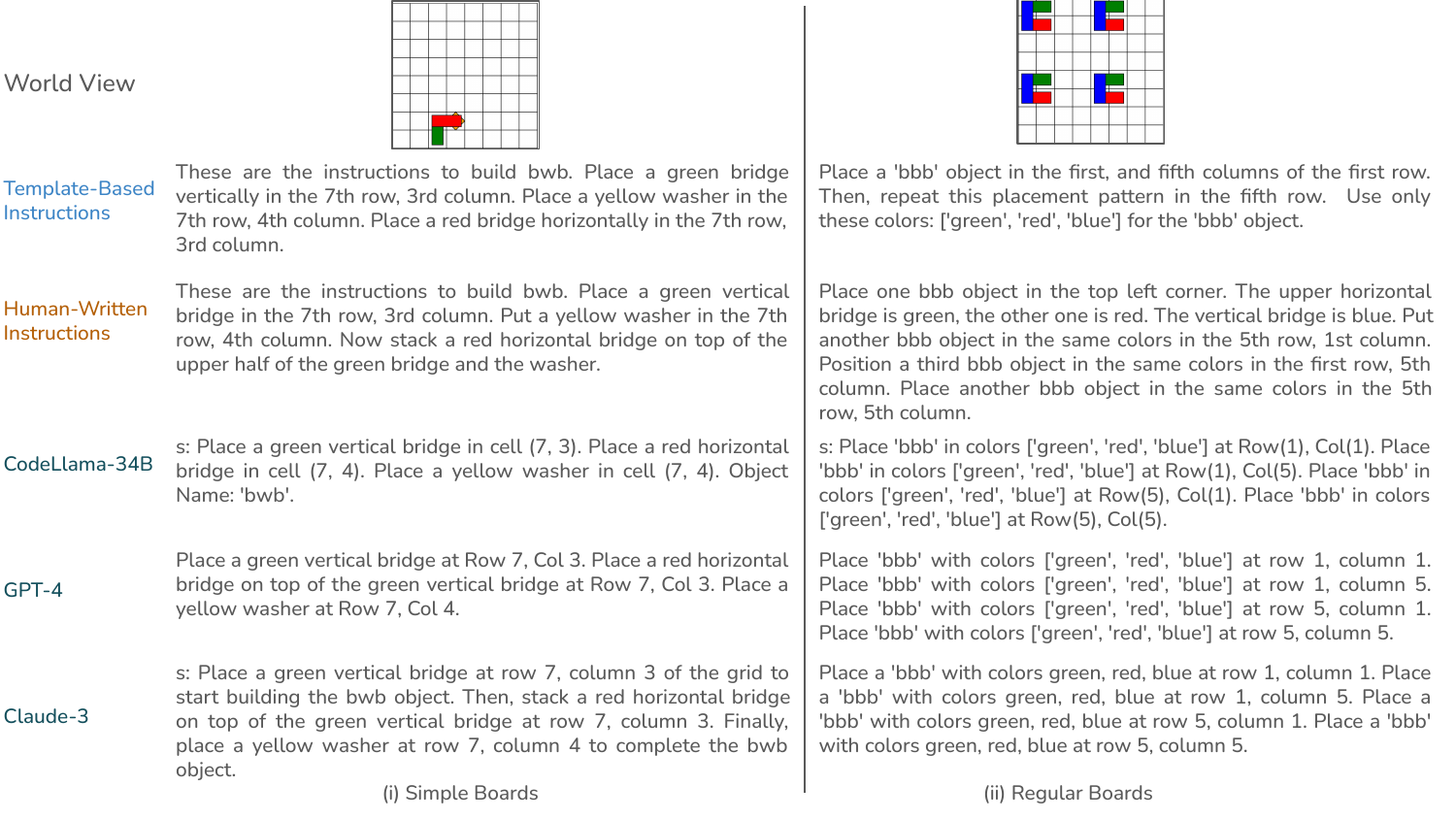}
  \caption {Three types of instructions pair with the \textit{simple} and \textit{regular} boards. Template-based instructions are generated using a template grammar. Human-author instructions are prepared by a human instructor. The remaining instructions are generated by three LLMs that describe the target.}
  \label{fig:modelgen_instructions}
\end{figure*}

At this step, we have pairs of target boards and code (in three variations) that generates them. Next, we add appropriate natural language instructions that verbalise the code and describe the target board, in three different ways. Figure~\ref{fig:modelgen_instructions} showcases the different types of instructions.

\paragraph{Template-based:} To generate natural language instructions automatically, we use a set of templates (see Appendix~\ref{sec:appendix_template_based}) that include grammar entries  defined in the \textit{Jinja2} format. This approach is inspired by work on data synthesis in other domains (e.g.\  \citep{DBLP:conf/aaai/RastogiZSGK20, DBLP:conf/sigdial/AksuLKC21}). These templates generate detailed and unambiguous instructions for reconstructing the target board. Each instruction sequence is available in both single-turn and multi-turn variants for all the splits.

\paragraph{Human-written:} In addition to template-based instructions, we curated human-written instructions to introduce natural and varied linguistic styles for the test set. Using the \textit{slurk}\citep{DBLP:conf/lrec/GotzePLDS22} framework, we implemented an interface to collect these human-authored instructions. We recruited an annotator whose task was to write instructions for reconstructing a given target board. Further details of the setup are provided in Appendix~\ref{sec:appendix-inst-human}. All human-written instructions consist of a single, complete message covering the entire board reconstruction, without being split into multiple steps.

\paragraph{Model-generated:} We employed an LLM to generate the required instructions to reconstruct the target board. The rationale behind using LLMs for this process includes their adaptability to a) generate instructions with varying nuances, b) maintain a standard and consistent format, and c) support multiple languages. Furthermore, experiments evaluating the effectiveness of such instructions for code generation offer opportunities for automation. Similar to human-written instructions, we generated these instructions for the target boards in the test set. Figure~\ref{fig:modelgen_instructions} showcases the instructions generated by three code-generation LLMs for reconstructing the board. Particulars on the input representation (ASCII representation of the target board is used in probing the model), prompt settings etc. are discussed in detail in Appendix~\ref{sec:appendix-inst-mg}. All LLM-generated instructions are single-turn.

\section{From Language to Robot Programs}
\label{sec:methodology}

At this point, we now have tuples consisting of a) code and b) natural language expressions, c) both of which describe the same target board (structure). With this setup, we can investigate to what extent LLMs can realise a function that takes NL expressions into code, with the meaning (the target structure) as invariant; where the function retrieval is possibly being helped by in-context learning from examples.

\subsection{Code Generation Task}
Given a natural language instruction describing a target board configuration, the goal is to generate Python code which, when executed (details on execution correctness evaluation are in Section~\ref{subsec:evalmetrics}), reconstructs the specified structure on a 2.5D grid. Each instruction may refer to component types, colors, positions, stacking rules, or higher-level patterns. The model is prompted in a few-shot setting. The generated code must adhere to spatial constraints, such as component placement, orientation, and stacking logic, and in some cases, should also employ reusable function definitions to represent repeated substructures. This task evaluates a model's ability to understand spatial language, generalize across structural variations, and translate abstract descriptions into executable symbolic programs.

We design our experiments to delve deeply into the generalizations~\citep{DBLP:journals/jair/HupkesDMB20} these models can make. We are interested in the following aspects of the process: 
a) \textit{Property Compositionality} - can models generalize from in-context examples that are similar yet vary in properties such as shape, color, and location? By curating examples that differ from the test instructions, we assess whether the model is merely copying or generalizing. b) \textit{Function Compositionality} - do the semantic understanding and pattern matching abilities of LLMs aid in generating higher-order code? and c) \textit{Function Repeatability} - can LLMs detect and reason how to repeat higher-order functions as per input instructions? This aspect tests LLMs’ on two fronts: understanding abstract instructions and optimizing code for repetitions. Such an approach is relevant in cobot programming, where repetitive tasks demand optimal handling.

\begin{figure*}
\centering
  \includegraphics[width=\textwidth]{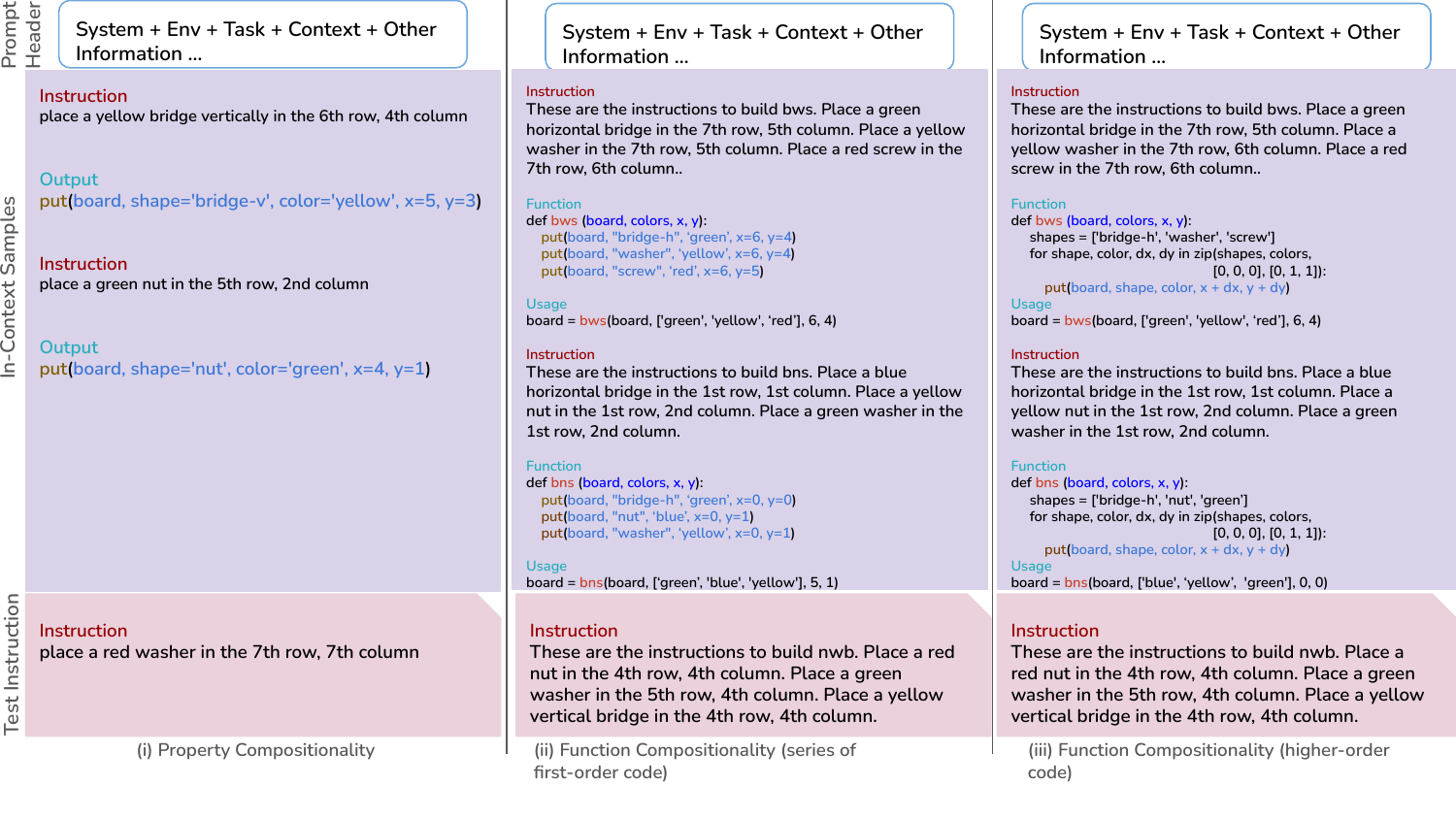}
  \caption {Illustration of prompt for evaluating property and function compositionality. \textbf{Left}: first-order code examples with varied component properties (name, color, location) for property compositionality testing. \textbf{Middle}: the construction of higher-order functions from first-order code snippets for function compositionality evaluation. \textbf{Right}: the construction of higher-order functions with optimal code for function compositionality evaluation.}
  \label{fig:incontext_samples_sb}
\end{figure*}

\section{Experimental Setup}
\label{sec:expsetup}

Following previously reported prompting approaches~\citep{DBLP:conf/nips/BrownMRSKDNSSAA20, DBLP:conf/corl/IchterBCFHHHIIJ22, DBLP:conf/icra/LiangHXXHIFZ23}, we constructed a multi-part prompt (see Figure~\ref{fig:incontext_samples_sb}), which we validated through an ablation study (see Figure~\ref{fig:prompt_basic}, Table~\ref{tab:prompt_structure} in Appendix~\ref{sec:appendix}). We used instruction-tuned code generation LLMs such as \textit{GPT-4} (version 1106-preview), \textit{CodeLlama-34B-instruct}~\citep{DBLP:journals/corr/abs-2308-12950} and \textit{Claude-3} (version \textit{opus}), with a \textit{temperature} of 0 and \textit{max\_new\_tokens} of 250.

The training split is used exclusively for in-context samples (see Figure~\ref{fig:incontext_samples_sb} in Appendix~\ref{sec:appendix-prompt-structure}). Our ablation study (detailed in Appendix~\ref{subsec:appendix_ablation}) indicated that using \textit{five} in-context examples yields the best results across various LLMs. We ensured the in-context examples did not share combinations of component types, and locations with the test instruction, which measures the LLMs responses for un-seen attributes and instructions. In real-world application, we lack control over in-context examples, and related examples are not necessarily detrimental. Thus, our results provide a lower bound on performance. The validation set is used for ablation studies, and the test set is used for evaluation. 

\begin{figure*}
  \includegraphics[width=\textwidth]{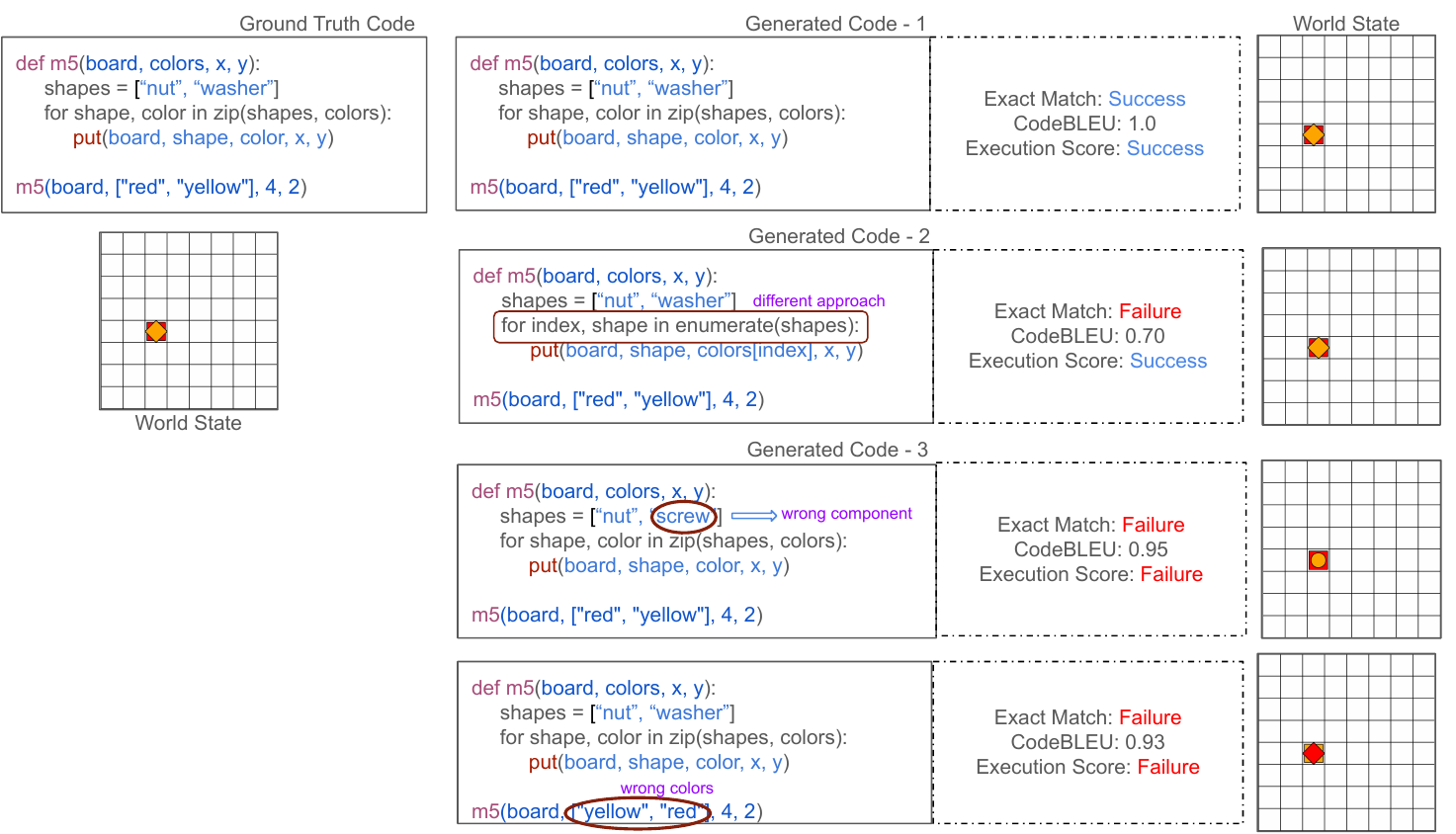}
  \caption {Overview of evaluation measurement for the generated code; Comparison of the generated code against ground truth using the metrics Exact Match (EM), Code BLEU score (CB), and Execution Success (ES); The ground truth code and its associated world state is shown on the left; Four examples of generated code are measured on the right; This highlights the strict criterion of EM, syntactic match criterion of CB and overall reconstruction accuracy of ES;}
  \label{fig:eval_metrics_overview}
\end{figure*}

\subsection{Evaluation Metrics}
\label{subsec:evalmetrics}

Compared to program synthesis and machine translation, our proposed task benefits from a known target configuration (i.e., a fully specified intended semantics), enabling a more nuanced evaluation of the generated output. First, we use exact match (EM) to compare the generated code with the gold-standard code (of the target structure) at the token level, assessing the LLMs' ability to produce semantically identical outputs. Second, the CodeBLEU score~\citep{DBLP:journals/corr/abs-2009-10297} evaluates the structural and functional quality of the generated code, testing the LLMs' capability to generate lexically and semantically correct code. Finally, we use execution success (ES) to compare the reconstructed structure with the gold-standard structure in terms of type, color, and location of elements to measure the LLMs' success in accurately reconstructing the target. This comprehensive evaluation (see Figure~\ref{fig:eval_metrics_overview}) examines semantic understanding, code quality, and precise execution abilities of the LLMs.

\section{Results and Analysis}
\label{sec:results}
As discussed in Section~\ref{sec:expsetup}, we investigate, how well LLMs benefit from in-context learning. Table~\ref{tab:res_overview} shows LLMs performance on various aspects of program synthesis and the impact of component arrangement complexity. Overall these models perform better on template-based instructions and the performance degraded for human-written and model-generated instructions. Detailed error analysis is provided in in Section~\ref{subsec:detail_error} and in Table~\ref{tab:error-info-details} (in Appendix).

%\noindent

\begin{table}
\centering
\caption{Code generation LLMs' performance across aspects of program synthesis. Evaluation includes atomic component placement (first-order code) and sequence arrangement (higher-order function generation), assessing the LLMs' ability to translate natural language input into executable code; EM - Exact Match, CB - Code BLEU, ES - Execution Success.
}
\label{tab:res_overview}
    \begin{subtable}{\textwidth}
        \caption{Template-based Instructions}    
        \begin{tabular}{|c|c|c|c|c|c|c|}
        \hline
        \begin{tabular}[c]{@{}c@{}}Board\\ Type\end{tabular} & \begin{tabular}[c]{@{}c@{}}Object\\ Type\end{tabular} & Task & Model & EM & CB & ES \\ \hline
        \multirow{9}{*}{Simple} & \multirow{9}{*}{Simple} & \multirow{3}{*}{\begin{tabular}[c]{@{}c@{}}Property Compositionality\end{tabular}} & CodeLlama & 0.97 & 0.99 & 0.97 \\ \cline{4-7} 
         &  &  & GPT-4 & 1.00 & 1.00 & \textbf{1.00} \\ \cline{4-7} 
         &  &  & Claude-3 & 1.00 & 1.00 & \textbf{1.00} \\ \cline{3-7} 
         &  & \multirow{3}{*}{\begin{tabular}[c]{@{}c@{}}Function Compositionality\\ (Using sequences of first-order code)\end{tabular}} & CodeLlama & 0 & 1.00 & 0.96 \\ \cline{4-7} 
         &  &  & GPT-4 & 0 & 1.00 & \textbf{1.00} \\ \cline{4-7} 
         &  &  & Claude-3 & 0 & 0.75 & 0.93 \\ \cline{3-7} 
         &  & \multirow{3}{*}{\begin{tabular}[c]{@{}c@{}}Function Compositionality\\ (Using optimal higher-order code)\end{tabular}} & CodeLlama & 0 & 0.95 & 0.52 \\ \cline{4-7} 
         &  &  & GPT-4 & 0 & 0.95 & 0.56 \\ \cline{4-7} 
         &  &  & Claude-3 & 0 & 0.97 & \textbf{0.87} \\ \hline
        \multirow{6}{*}{Regular} & \multirow{3}{*}{Simple} & \multirow{6}{*}{\begin{tabular}[c]{@{}c@{}}Function Repeatability\end{tabular}} & CodeLlama & 0 & 0.99 & 0.75 \\ \cline{4-7} 
         &  &  & GPT-4 & 0 & 0.96 & \textbf{1.00} \\ \cline{4-7} 
         &  &  & Claude-3 & 0 & 0.95 & 0.86 \\ \cline{2-2} \cline{4-7} 
         & \multirow{3}{*}{Complex} &  & CodeLlama & 0 & 0.22 & 0.09 \\ \cline{4-7} 
         &  &  & GPT-4 & 0 & 0.49 & \textbf{0.30} \\ \cline{4-7} 
         &  &  & Claude-3 & 0 & 0.51 & 0.10 \\ \hline
        \end{tabular}
    \end{subtable}%
    \vspace{1em}
    \begin{subtable}{\textwidth}
        \caption{Human-written Instructions}    
        \begin{tabular}{|c|c|c|c|c|c|c|}
        \hline
        \begin{tabular}[c]{@{}c@{}}Board\\ Type\end{tabular} & \begin{tabular}[c]{@{}c@{}}Object\\ Type\end{tabular} & Task & Model & EM & CB & ES \\ \hline
        \multirow{3}{*}{Simple} & \multirow{3}{*}{Simple} & \multirow{3}{*}{\begin{tabular}[c]{@{}c@{}}Function Compositionality\\ (Using optimal higher-order code)\end{tabular}} & CodeLlama & 0 & 0.95 & 0.28 \\ \cline{4-7} 
         &  &  & GPT-4 & 0 & 1.00 & 0.17 \\ \cline{4-7} 
         &  &  & Claude-3 & 0 & 0.79 & \textbf{0.43} \\ \hline
        \multirow{6}{*}{Regular} & \multirow{3}{*}{Simple} & \multirow{6}{*}{\begin{tabular}[c]{@{}c@{}}Function Repeatability\end{tabular}} & CodeLlama & 0 & 0.34 & 0.07 \\ \cline{4-7} 
         &  &  & GPT-4 & 0 & 0.49 & \textbf{0.31} \\ \cline{4-7} 
         &  &  & Claude-3 & 0 & 0.69 & 0.25 \\ \cline{2-2} \cline{4-7} 
         & \multirow{3}{*}{Complex} &  & CodeLlama & 0 & 0.08 & 0.07 \\ \cline{4-7} 
         &  &  & GPT-4 & 0 & 0.09 & \textbf{0.28} \\ \cline{4-7} 
         &  &  & Claude-3 & 0 & 0.11 & 0.04 \\ \hline
        \end{tabular}
    \end{subtable}%
    \vspace{1em}
    \begin{subtable}{\textwidth}
        \caption{Model-generated Instructions}    
        \begin{tabular}{|c|c|c|c|c|c|c|}
        \hline
        \begin{tabular}[c]{@{}c@{}}Board\\ Type\end{tabular} & \begin{tabular}[c]{@{}c@{}}Object\\ Type\end{tabular} & Task & Model & EM & CB & ES \\ \hline
        \multirow{3}{*}{Simple} & \multirow{3}{*}{Simple} & \multirow{3}{*}{\begin{tabular}[c]{@{}c@{}}Function Compositionality\\ (Using optimal higher-order code)\end{tabular}} & CodeLlama & 0 & 0.64 & 0.04 \\ \cline{4-7} 
         &  &  & GPT-4 & 0 & 0.85 & 0.22 \\ \cline{4-7} 
         &  &  & Claude-3 & 0 & 0.92 & \textbf{0.23} \\ \hline
        \multirow{3}{*}{Regular} & \multirow{3}{*}{Simple} & \multirow{3}{*}{\begin{tabular}[c]{@{}c@{}}Function Repeatability\end{tabular}} & CodeLlama & 0 & 0.19 & 0 \\ \cline{4-7} 
         &  &  & GPT-4 & 0 & 0.19 & \textbf{0.13} \\ \cline{4-7} 
         &  &  & Claude-3 & 0 & 0.09 & 0.33 \\ \hline
        \end{tabular}
    \end{subtable}%  
\end{table}

\paragraph{Property Compositionality:} We assess if LLMs can synthesize input instructions into first-order code for unseen instructions and attributes. High scores across all measurements (EM, CB, and ES) show that LLMs are good at interpreting the instruction, extracting the attributes such as color, shape, and location, mapping the locations to the current environment (an 8x8 grid), and generating semantically identical code. This demonstrates that domain specific first-order code generation is achievable with few-shot prompting in both open-source and closed API-based LLMs.

\paragraph{Function Compositionality:} We further investigate if LLMs show high performance on functional compositionality, assessing their ability to generate higher-order functions by abstracting the functionality into reusable programs. Zero scores for EM metric indicate that the generated code is not semantically identical to the ground truth and may have formatting discrepancies. Lower ES scores indicate that LLM-generated code struggles to accurately reconstruct the target board, highlighting the challenges in complex scenarios.

\begin{figure*}[t]
\centering
  \includegraphics[width=0.92\textwidth]{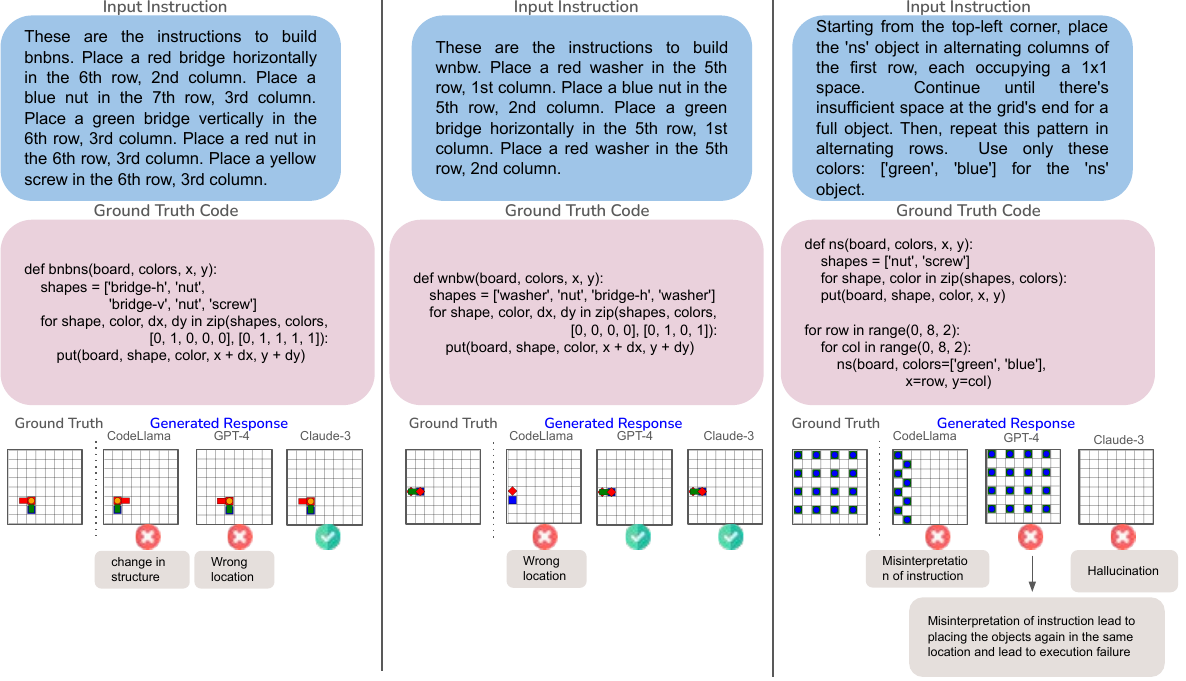}
  \caption {Execution response of the code generated by models. It highlights the types of errors encountered, including incorrect object placement, structural distortions, and hallucinations.}
  \label{fig:qualitative_analysis}
  \vspace{-0.20cm}
\end{figure*}

Manual analysis of model responses (as illustrated in Figure~\ref{fig:qualitative_analysis}), revealed factors reducing the execution scores. CodeLlama-34B had location-related errors in 89\% of template-based and 95\% of human-written instructions, causing depth mismatches and incorrect placements. GPT-4 and Claude-3 also had location-related errors, with all errors in GPT-4 and 95\% in Claude-3 due to incorrect locations. This indicates that in-context learning is less-effective for complex function generation and requires further exploration.

\paragraph{Function Repeatability:} We continue our analysis of LLMs abilities in generating loops, including nested ones for function repeatability. Zero EM scores result from strict matching criteria. While models perform well with simple objects, they struggle with complex objects, due to intricate patterns requiring overlap management. GPT-4 was effective when objects don't overlap. Claude-3 often hallucinated, with about 79\% of errors linked to this problem. CodeLlama had about 92\% of errors resulting from failure to consider overlaps. These results highlight the challenges of learning repetitions and the difficulty of incorporating this ability through in-context learning alone.

Our comprehensive analysis reveals that in-context learning is beneficial for LLMs in generating first-order code and handling simple loops, but underscore the need for more advanced techniques in generating code for complex and nested functions.

\subsection{Detailed Error Analysis}
\label{subsec:detail_error}
The detailed error analysis focuses on issues that lead to reduced execution scores for the LLM-generated response, as showcased in Table~\ref{tab:error-info-details} (in Appendix). These errors are broadly classified into two categories: board placement errors and element mismatch errors. Board placement errors refer to instances where the generated response cannot be executed, while element mismatch errors occur when the execution has discrepancies in location, color, or component.

The following list highlights the different scenarios where board placement errors can occur.
\begin{enumerate}
    \item \textbf{Syntax Error:} Incorrect indentation or inclusion of non-Python code, such as hallucinations.
    \item \textbf{Key Error:} Use of colors or components not supported in the environment.
    \item \textbf{Name Error:} References to unsupported functions or random function calls.
    \item \textbf{Value Error:} Placement of bridges at grid boundaries, specifically the 7th row for vertical bridges and the 7th column for horizontal bridges.
    \item \textbf{Dimensions Mismatch Error:} Locations specified outside the grid boundaries, exceeding an 8x8 grid.
    \item \textbf{Depth Mismatch Error:} Stacking of horizontal or vertical bridges without proper depth alignment.
    \item \textbf{Bridge Placement Error:} Stacking of bridges at height-3 or higher.
    \item \textbf{Same Shape Stacking Error}, \textbf{Same Shape At Alternate Levels Error}: Occur when identical shapes are stacked incorrectly.
    \item  \textbf{Not On Top Of Screw Error}: Placement of components on top of a \textit{screw}.
\end{enumerate}

It is important to note that the target structures used in the test split do not contain any invalid scenarios that may cause these errors. Ideally, if the information is accurately extracted from the instructions, none of these errors would occur. The quantitative analysis described in Table~\ref{tab:res_overview} demonstrate that models struggle to interpret the instructions and extract relevant information, often making mistakes in associating the correct order, color, count, and location of components.

\section{ Conclusion and Limitations}
\label{sec:conclusion}
This paper explores the program synthesis capabilities of code-generating LLMs, whose understanding can significantly enhance conversational programming for cobots. To this end, we developed a rapid prototyping simulator environment that evaluates these LLMs using specific input instructions, simulate their execution, and analyzes the output. This approach establishes a strong baseline for future studies. Our findings reveal promising results for template-based instructions that generate first-order code and functions composed of sequences of first-order code. However, performance significantly declines with optimal higher-order functions and complex patterns that require reasoning about object overlap in repetitions. Such a finding indicates a pressing need for further investigation. Similarly the LLMs performance for human-written and model-generated instructions is considerably lower compared to template-based instructions, calls for a detailed exploration. In future work, we plan to expand the setup to handle more complex sequence arrangements added with dialogue management. Additionally, we will explore fine-tuning the code generation LLMs to see if it improves their overall performance. 

Though LLMs have shown promising performance for the 2.5D building task, there are limitations. The seeds for target board generation focus only on simple objects; we plan to include complex objects to better assess capabilities. The template-based instructions were designed to be clear and unambiguous, establishing a performance baseline. Future work will include environmental variability, collaborative scenarios, and dialogue management to enhance real-world simulation. Despite careful prompt curation, models sometimes produce hallucinations. Refining prompts based on feedback from execution errors can help. Instruction-tuned LLMs for code generation may struggle with multi-turn conversations, and applying this work to real-world assembly tasks will require adherence to safety principles.

\begin{acknowledgments}
The work reported here has been funded by the Bundesministerium für Bildung und Forschung (BMBF, German Federal Ministry of Research), project "COCOBOTS" (01IS21102A) and Deutsche Forschungsgemeinschaft (DFG, German Research Foundation) grant 423217434 (“RECOLAGE”). We thank the reviewers for their helpful feedback.
\end{acknowledgments}

%%
%% Define the bibliography file to be used
\bibliography{sample-1col_cameraready}

\appendix
\label{sec:appendix}
\section{Simulated Environment}
\label{sec:appendix_simul_env}
As mentioned in Section~\ref{sec:taskenv}, our proposed simulated environment uses a 2.5D grid of size 8x8, where each cell of the grid can hold a component. Any specific arrangement of components on the grid should adhere to the rules such as a) components of the same type or color cannot be stacked on each other and b) components can be placed atop one another as long as the depths (the number of cells occupied by the components) match, with the sole exception that no components can be placed on a \textit{screw}. These rules allow us to construct component arrangements that resemble real-world style assemblies. Each such arrangement, where the components are connected, in such a way that electricity flows throw them is referred as an object. These objects are categorized as simple and complex based on their configurations. \textit{Simple objects} are basic configurations of up to five components and no more than three stacks. 

\section{Template-Based Instructions}
\label{sec:appendix_template_based}
\paragraph{Instruction Templates:} We employed a standard template grammar for generating instructions based on templates. Simple boards are organized into two categories: a) multi-turn and b) single-turn. As illustrated in Figure~\ref{fig:sb_so_multiturn}, multi-turn instructions consist of sequential steps described over 'n' turns. In contrast, single-turn instructions, depicted in Figure~\ref{fig:sb_so_singleturn}, compile all steps necessary to construct a target structure within a single turn. For regular boards, the template grammar applied to simple and complex objects is presented in Figure~\ref{fig:instruction_templates_rb_so} and Figure~\ref{fig:instruction_templates_rb_co}, respectively. Both templates articulate the specific arrangements utilized. Moreover, the complex object templates include placeholders for each object's space, which is critical for the instruction follower when generating the construction code.

\paragraph{Code Templates:} Figures~\ref{fig:code_templates_1}-\ref{fig:code_templates_7} showcase the Jinja2 templates used for generating the code for higher-order functions for the manually curated code seeds (see Section~\ref{subsubsec:boardgen}). The placeholders for shapes, colors, and locations are replaced with appropriate values during the board generation process.

\section{Human-Written Instructions}
\label{sec:appendix-inst-human}
As mentioned in Section~\ref{subsubsec:boardgen}, we curated human instructions for the test set (see Table~\ref{tab:dataset_stats}) to measure how well the LLMs grasp human abstractions and communication styles for the code generation. We developed a bot using \textit{slurk}~\citep{DBLP:conf/lrec/GotzePLDS22} interface, which features a target board on the left and a text input area on the right for writing the instructions to reconstruct it. A student assistant from our department was recruited for this task, which took approximately 20hours to complete.

\section{Model-Generated Instructions}
\label{sec:appendix-inst-mg}
\begin{figure}[tb]
\centering
    \begin{prompt}
\\
\textbf{System Info}
\\\\
You are an expert annotator who generates sequential instructions for populating a grid with the given shapes.
\\\\
\textbf{Environment Info}
\\\\
The environment is an 8x8 grid allowing shape placement and stacking. A shape can be placed in any cell, while stacking involves adding multiple shapes to the same cell, increasing its depth. Shapes typically occupy a single cell, except for the "bridge," which spans two cells and requires two other shapes for stacking. Horizontal bridges span adjacent columns (left and right), and vertical ones span consecutive rows (top and bottom). Stacking is only possible if the shapes have matching depths.
\\\\
In the grid, columns align with the x-axis and rows with the y-axis. The cell in the top-left corner is the first row and first column, corresponding to row and column values of 1, 1. Similarly, the top-right corner cell is the first row and eighth column, with row and column values of 1, 8.
\\\\
Some of the cells in the grid are filled with shapes, and the current status of the grid is labeled under `Current Grid Status'. If multiple shapes are placed in the same cell, they are mentioned in the order from bottom to top. All the shapes combined are referred to as an `object', and the name of the object is labeled under `Object Name'. Each filled cell in the grid contains a list of tuples, where each tuple indicates the name of the shape and its color. Empty cells are indicated by `\includegraphics[width=1em]{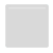}'.
\\\\
The elaboration about the grid is labeled under 'Grid Explanation'.
\\\\
\textbf{Task Info}
\\\\
Your task is to respond with the sequential instructions under the label Instruction followed by a newline.

Generate the instructions to fill the grid with given shapes, listing all steps in a continuous format without numbering or bullet points. Also ensure to mention the object name in the instructions. Assume the grid starts empty and only describe actions for placing shapes. The order of colors, x, y matters, as these are assigned to the shapes in the same sequence.
\\\\
\textbf{Other Info}
\\\\
Do not generate any other text/explanations.

Lets begin
\\\\
\$CURRENT\_GRID\_STATUS
\\\\
\$GRID\_EXPLANATION

\end{prompt}
\caption{prompt template for probing LLM to generate instruction for simple boards}
    \label{fig:prompt_mg_sb}
\end{figure}

\begin{figure}[tb]
    \begin{prompt}
\\
\textbf{System Info}
\\\\
You are an expert annotator who generates sequential instructions for populating a grid with the given shapes.
\\\\
\textbf{Environment Info}
\\\\
The environment is an 8x8 grid allowing shape placement and stacking. A shape can be placed in any cell, while stacking involves adding multiple shapes to the same cell, increasing its depth. Shapes typically occupy a single cell, except for the "bridge," which spans two cells and requires two other shapes for stacking. Horizontal bridges span adjacent columns (left and right), and vertical ones span consecutive rows (top and bottom). Stacking is only possible if the shapes have matching depths.
\\\\
In the grid, columns align with the x-axis and rows with the y-axis. The cell in the top-left corner is the first row and first column, corresponding to row and column values of 1, 1. Similarly, the top-right corner cell is the first row and eighth column, with row and column values of 1, 8.
\\\\
Some of the cells in the grid are filled with objects, and the current status of the grid is labeled under `Current Grid Status'. Each filled cell in the grid contains a list of tuples, where each tuple indicates the name of the object and its colors. Empty cells are indicated by `\includegraphics[width=1em]{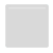}'.
\\\\
The elaboration about the grid is labeled under 'Grid Explanation'.
\\\\
\textbf{Task Info}
\\\\
Your task is to respond with the sequential instructions under the label Instruction followed by a newline.

Generate the instructions to fill the grid with the given object, in a continuous format without numbering or bullet points. Assume the grid starts empty and only describe actions for placing the object. The order of colors, x, y matters, as these are assigned to the object in the same sequence.
\\\\
\textbf{Other Info}
\\\\
Do not generate any other text/explanations.

Lets begin
\\\\
\$CURRENT\_GRID\_STATUS
\\\\
\$GRID\_EXPLANATION

\end{prompt}
\caption{prompt template for probing LLM to generate instruction for regular boards }
    \label{fig:prompt_mg_rb}
\end{figure}

As described in Section~\ref{subsubsec:instructiongen}, LLMs are used to describe the target structure. These descriptions are later used as part of the component assembly task for generating the associated code. Using the \textit{clembench} framework~\citep{DBLP:conf/emnlp/ChalamalasettiG23}, we setup the process of describing the target structure as a \textit{clemgame}. In this game, the game master probes the player (a code generation LLM) with the grid details represented in a textual format as shown in the Figure~\ref{fig:prompt_mg_sb} and Figure~\ref{fig:prompt_mg_rb} using zero-shot prompting techniques~\citep{DBLP:conf/nips/BrownMRSKDNSSAA20, DBLP:conf/corl/IchterBCFHHHIIJ22, DBLP:conf/icra/LiangHXXHIFZ23} to generate the textual instructions. 

\section{Prompt Structure}
\label{sec:appendix-prompt-structure}
The principal objective of the proposed work is to investigate the capability of the LLM in capturing the procedural steps involved in arranging components in a specific sequence, generating an abstract code representation that is both execution-ready and applicable within the simulation environment. To achieve execution readiness, the LLM must extract requisite details from instructions and transform them into an intended structural format expressed through context information and in-context samples. Building on this, we construct a multi-part prompt (shown in Figure~\ref{fig:prompt_basic}) to probe the LLM. Each section of this prompt is designed to convey distinct pieces of information, clearly defining the task's goals and expected response for the LLM.

System Information guides the LLM toward the overall desired outcome.  
The simulation setup uses distinct names for components based on their orientation properties, which is crucial for the LLM to identify and utilize the correct names accurately. In our simulation setup, columns increase along the x-axis, and rows along the y-axis. It is important to note that the numbering begins at the top, unlike the conventional approach, which starts at the bottom. This places the top-left corner as the first row and first column. Understanding this unique orientation is essential for the LLM to translate spatial information from instructions into code correctly. Therefore, these specifics are included as part of the environment information.

Following this, the context information specifies the functions available in the environment, eliminating the need for the LLM to generate new code for these functions. After the context information, task information provides the labels the LLM should use in its responses. This assists the parser in identifying and extracting the relevant responses. In-context samples, which come next, illustrate the types of instructions and associated actions the LLM encounters. In adherence to prompt engineering best practices, we have also included explicit instructions to prevent the generation of explanations or instructions other than the responses labeled as specified in the task information. All the experiments for property compositionality, function compositionality, and function repeatability follow the same prompt structure but differ in terms of the in-context samples they use. The structure of in-context samples used for each of these experiments is shown in Figure~\ref{fig:incontext_samples_sb}).

\subsection{Ablation Study}
\label{subsec:appendix_ablation}

We conducted an ablation study to investigate the impact of various parts of the prompt on overall task performance. We used the validation split of the dataset to perform the study. Table~\ref{tab:prompt_structure} provides a comprehensive overview of the ablation study results, presenting the performance metrics for different prompt configurations. The analysis focuses on determining optimal prompts for capturing procedural steps and generating execution-ready, simulation-applicable code representations. Each row in the table corresponds to a specific prompt structure element, and the columns include relevant performance metrics, such as the execution score for each code LLM, to quantify the impact of these elements on the overall effectiveness in code generation. The results demonstrate that the structure shown in Figure~\ref{fig:prompt_basic} is optimal for all the tasks and LLMs used in our experiments. Overall, the prompting structure lacking task information and in-context samples deteriorated the performance, while adding environment and context information improved the performance slightly. System details helped guiding the desired outcome from LLM. Therefore, we use a prompt structure comprising \textit{system}, \textit{environment}, \textit{task}, \textit{context}, \textit{five in-context samples} and \textit{other information} for the remaining experiments.

\subsection{Selection of In-context Samples}
\label{subsec:appendix_icsamples}
\begin{figure*}
  \includegraphics[width=\textwidth]{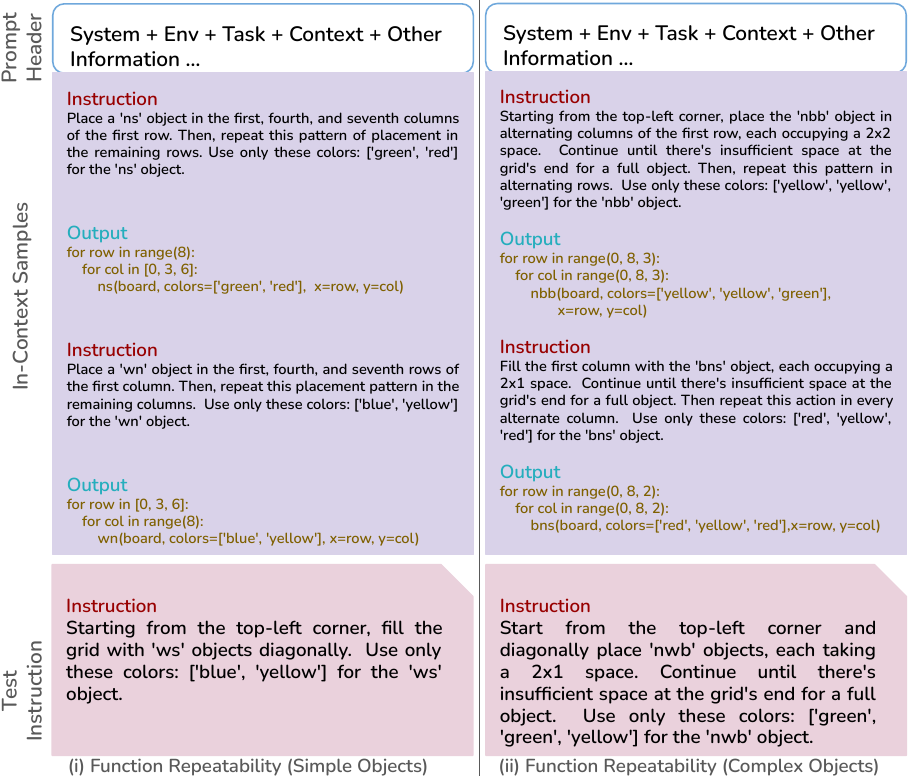}
  \caption {In-context samples for regular boards.}
  \label{fig:incontext_samples_rb}
\end{figure*}

The training split is used to select in-context samples for prompts. Our ablation study described above shows that using \textit{five} in-context samples yields the best results across various LLMs. We dynamically prepare examples for each sample in the test split, ensuring they do not share the test instruction’s component combination to avoid overlap and bias. From this pool, five random samples are chosen for each test sample, maintaining their uniqueness and relevance.
Figure~\ref{fig:incontext_samples_sb} and Figure~\ref{fig:incontext_samples_rb} displays examples illustrating \textit{simple} boards and \textit{regular} boards highlighting the tailored in-context samples for each scenario.

\section{Evaluation Metrics}
\label{sec:appendix_eval_metrics}
\begin{figure*}
  \includegraphics[width=\textwidth]{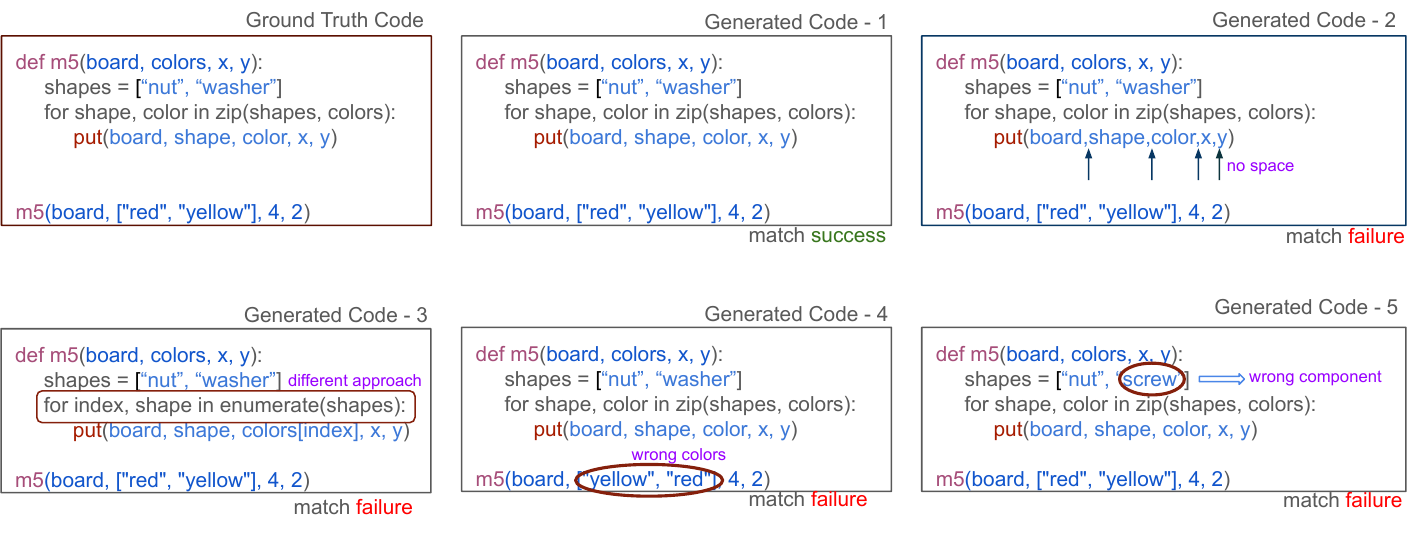}
  \caption {Overview of Exact Match measurement for the generated code; The ground truth code and its associated world state are shown on the left. Although exact match is able to detect mismatches in attributes (colors and components), it fails to detect the functional correctness of code with the same semantic logic.}
  \label{fig:em_metric_overview}
\end{figure*}

\begin{figure*}
  \includegraphics[width=\textwidth]{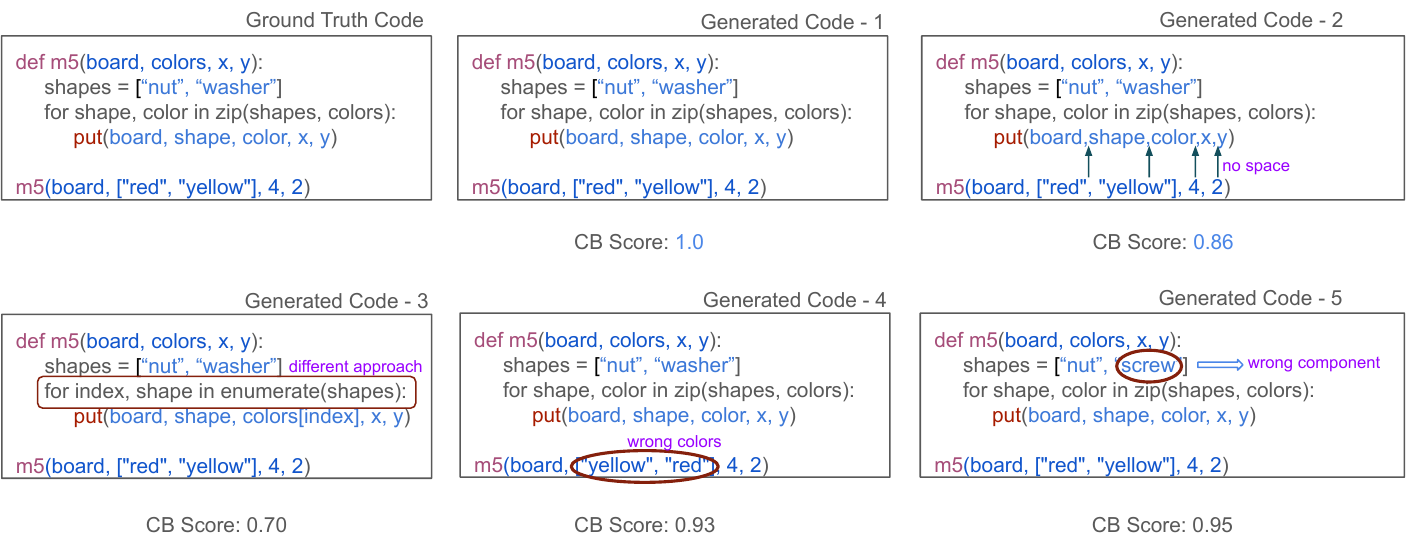}
  \caption {Overview of Code BLEU score measurement for the generated code; The ground truth code and its associated world state are shown on the left. The Code BLEU score is able to detect the syntactic correctness of the code (different styles), and also have high scores for the code where the similar style is followed but uses in-correct attributes. Since the values of attributes has no significance in Code BLEU score computation, it fails to detect mismatches in the attributes and cannot be relied upon completely for such reconstruction tasks.}
  \label{fig:cb_metric_overview}
\end{figure*}

\begin{figure*}
  \includegraphics[width=\textwidth]{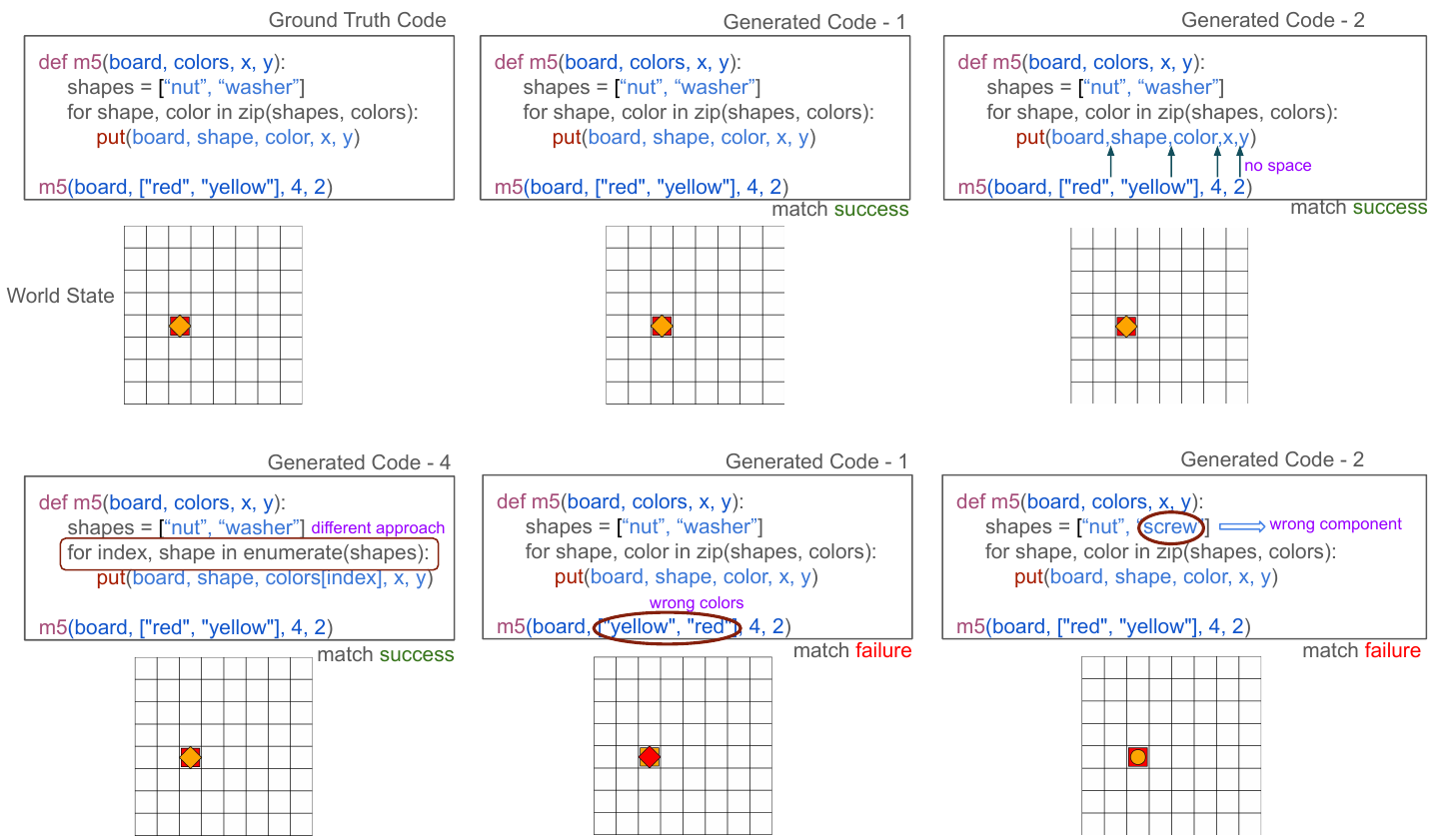}
  \caption {Overview of Execution Success measurement for the generated code; The ground truth code and its associated world state are shown on the left. Execution success compares the resultant 2.5D grid for its cell values (type, color and location of each component) against the ground truth grid state. As a result, the score reflects functional correctness along with mismatches in the attributes.}
  \label{fig:es_metric_overview}
\end{figure*}

We have followed three different metrics to evaluate the LLM generated code. First, we use exact matching to compare the generated code with the ground truth code. Figure~\ref{fig:em_metric_overview} showcases how this measurement uses a strict equality criterion for matching. The generated code (examples indicated by generated code-1, 2, 3, 4, 5 in Figure~\ref{fig:em_metric_overview}) is compared with the ground truth string and any mismatches (formatting, logical, or functional) in the code are treated as a failure. This score is useful to measure the perfect accuracy, but is inefficient in capturing semantic logic. Second, we use Code BLEU score to measure the syntactic accuracy of the generated code. The Code BLEU score~\citep{DBLP:journals/corr/abs-2009-10297} (CB) uses weighted combination of n-gram match, syntactic AST match and semantic data-flow match. We use the python package, codebleu~\footnote{\url{https://github.com/k4black/codebleu}} to compute the score. \textit{calc\_codebleu() API} takes the ground truth code and generated code as inputs and returns the overall Code BLEU score. Figure~\ref{fig:cb_metric_overview} shows how the CB score varies for differences in the generated code w.r.t the ground truth code. This score is useful to measure the syntactic correctness and is unreliable in capturing the overall accuracy. Lastly, we use execution success to measure how accurately the target board is reconstructed.  Figure~\ref{fig:es_metric_overview} demonstrates how the current grid state changes after the execution of the generated code. The example generated code-1, 2, and 3 share the same world state (row:5, column: 3 has two components: nut:red and washer:yellow) as the ground truth and are treated as successes. Although the example generated code-4 shares similar location and component types with the ground truth code, they differ in the colors associated with each component type (nut:yellow and washer:red) and is marked as a failure. Similarly, the example generated code-5 has a different component (screw:red) at the location(row:5, column: 3) and is marked as a failure. Thus, the execution success score captures semantic identicalness, functional correctness, making it reliable for target board reconstruction tasks.

\begin{figure*}
  \centering
  \begin{subfigure}[b]{0.49\textwidth}
    \centering
    \begin{prompt}
\\
\textbf{System Info}
\\\\
You are a helpful assistant who is designed to interpret and translate natural language instructions into python executable code snippets.
\\\\
\textbf{Environment Info}
\\\\
The environment is an 8x8 grid allowing shape placement and stacking. A shape can be placed in any cell, while stacking involves adding multiple shapes to the same cell, increasing its depth. Shapes typically occupy a single cell, except for the "bridge," which spans two cells and requires two other shapes for stacking. Horizontal bridges span adjacent columns (left and right), and vertical ones span consecutive rows (top and bottom). Stacking is only possible if the shapes have matching depths.
\\\\
In the grid, columns align with the x-axis and rows with the y-axis. Python indexing is used to identify each cell. The cell in the top-left corner is in the first row and first column, corresponding to x and y values of 0, 0. Similarly, the top-right corner cell is in the first row and eighth column, with x and y values of 0, 7.
\\\\
- Use the shape name 'bridge-h' if a bridge is placed horizontally
- Use the shape name 'bridge-v' if a bridge is placed vertically
\\\\
\textbf{Context Info}
\\\\
The following functions are already defined; therefore, do not generate additional code for it
\\\\
- Use `put(board: np.ndarray, shape: string, color: string, x: int, y: int) to place a shape on the board
\end{prompt}
\caption{base structure of prompt}
    \label{fig:prompt_basic}
  \end{subfigure}
  \begin{subfigure}[b]{0.46\textwidth}
    \centering
\begin{prompt}
\\
\textbf{TASK INFO}
\\\\
The TASK INFO varies for each of the sub-tasks (property compositionality and function compositionality).
\\\\
For each instruction labeled \$INSTRUCTION\_LABEL please respond with code under the label \$OUTPUT\_LABEL followed by a newline.
\\\\
\textbf{INCONTEXT\_SAMPLES}
\\\\
\textit{The INCONTEXT\_SAMPLES varies for each of the sub-tasks (property compositionality and function compositionality). Refer to }Figure~\ref{fig:incontext_samples_sb} and Figure~\ref{fig:incontext_samples_rb}
\\\\
\textbf{OTHER DETAILS}
\\\\
Do not generate any other text/explanations.
\\\\
Ensure the response can be executed by Python `exec()`, e.g.: no trailing commas, no periods, etc.

Lets begin
\end{prompt}
\caption{task and in-context samples}
    \label{fig:prompt_task_incontext_basic}
  \end{subfigure}  
  \caption{Prompt templates used for code generation for the tasks of Property Compositionality and Function Compositionality. The system information specifies system level behavior, the environment information indicates the environment details of the simulation framework, the context information describes the available functions that can be reused, task information indicates the specific response format to follow based on task type (differs for property compositionality and function compositionality)}
  \label{fig:prompt_template}
\end{figure*}


\begin{figure*}
\begin{prompt}
\\
\textbf{CURRENT\_GRID\_STATUS}
\\\\
CURRENT\_GRID\_STATUS varies for each of the target boards
\\\\
\newline
[`\includegraphics[width=1em]{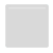}', `\includegraphics[width=1em]{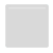}', `\includegraphics[width=1em]{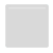}', `\includegraphics[width=1em]{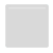}', `\includegraphics[width=1em]{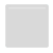}', `\includegraphics[width=1em]{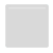}', `\includegraphics[width=1em]{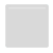}', `\includegraphics[width=1em]{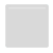}']
\newline
[`\includegraphics[width=1em]{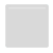}', `\includegraphics[width=1em]{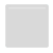}', `\includegraphics[width=1em]{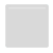}', `\includegraphics[width=1em]{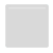}', `\includegraphics[width=1em]{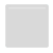}', `\includegraphics[width=1em]{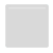}', `\includegraphics[width=1em]{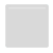}', `\includegraphics[width=1em]{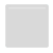}'] 
\newline
[`\includegraphics[width=1em]{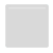}', `\includegraphics[width=1em]{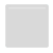}', `\includegraphics[width=1em]{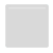}', `\includegraphics[width=1em]{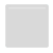}', `\includegraphics[width=1em]{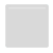}', `\includegraphics[width=1em]{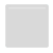}', `\includegraphics[width=1em]{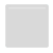}', `\includegraphics[width=1em]{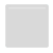}'] 
\newline
[`\includegraphics[width=1em]{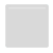}', `\includegraphics[width=1em]{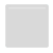}', `\includegraphics[width=1em]{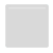}', `\includegraphics[width=1em]{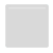}', `\includegraphics[width=1em]{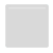}', `\includegraphics[width=1em]{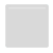}', `\includegraphics[width=1em]{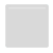}', `\includegraphics[width=1em]{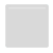}'] 
\newline
[`\includegraphics[width=1em]{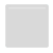}', `\includegraphics[width=1em]{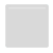}', `\includegraphics[width=1em]{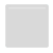}', `\includegraphics[width=1em]{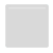}', `\includegraphics[width=1em]{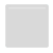}', `\includegraphics[width=1em]{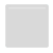}', `\includegraphics[width=1em]{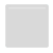}', `\includegraphics[width=1em]{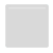}'] 
\newline
[`\includegraphics[width=1em]{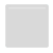}', `\includegraphics[width=1em]{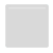}', `\includegraphics[width=1em]{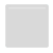}', `\includegraphics[width=1em]{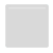}', `\includegraphics[width=1em]{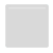}', `\includegraphics[width=1em]{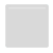}', `\includegraphics[width=1em]{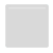}', `\includegraphics[width=1em]{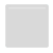}'] 
\newline
[`\includegraphics[width=1em]{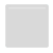}', `\includegraphics[width=1em]{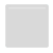}', [(`washer', `red'), (`screw', `blue')], `\includegraphics[width=1em]{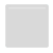}', `\includegraphics[width=1em]{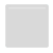}', `\includegraphics[width=1em]{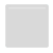}', `\includegraphics[width=1em]{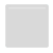}', `\includegraphics[width=1em]{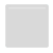}'] 
\newline
[`\includegraphics[width=1em]{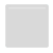}', `\includegraphics[width=1em]{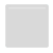}', `\includegraphics[width=1em]{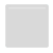}', `\includegraphics[width=1em]{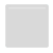}', `\includegraphics[width=1em]{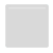}', `\includegraphics[width=1em]{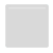}', `\includegraphics[width=1em]{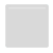}', `\includegraphics[width=1em]{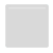}']
\\\\
\textbf{Object Name}
`ws'.
\\\\
\textbf{Grid Explanation}:
\\\\
Row(7), Col(3) contains red washer, blue screw.
\\\\
\end{prompt}
\caption{Overview of grid explanation in the prompt for describing a target structure}
\label{fig:prompt_template_grid_explanation}
\end{figure*}
\begin{table*}
\small
\centering
    \caption{Overview of ablation study experiments for the optimal number of in-context samples and the prompt structure}
    \label{tab:ablationstudy}
    \begin{subtable}[t]{\textwidth}
        \caption{Optimal number of in-context samples for the property compositionality task}
        \label{subtab:1}    
\begin{tabular}{|c|ccccccccccc|}
\hline
\multirow{2}{*}{Model} & \multicolumn{11}{c|}{Number of In-Context Samples} \\ \cline{2-12} 
 & \multicolumn{1}{c|}{0} & \multicolumn{1}{c|}{1} & \multicolumn{1}{c|}{2} & \multicolumn{1}{c|}{3} & \multicolumn{1}{c|}{4} & \multicolumn{1}{c|}{5} & \multicolumn{1}{c|}{6} & \multicolumn{1}{c|}{7} & \multicolumn{1}{c|}{8} & \multicolumn{1}{c|}{9} & 10 \\ \hline
CodeLlama-7B & \multicolumn{1}{c|}{0.53} & \multicolumn{1}{c|}{0.90} & \multicolumn{1}{c|}{0.92} & \multicolumn{1}{c|}{0.95} & \multicolumn{1}{c|}{0.95} & \multicolumn{1}{c|}{0.95} & \multicolumn{1}{c|}{0.95} & \multicolumn{1}{c|}{0.97} & \multicolumn{1}{c|}{0.98} & \multicolumn{1}{c|}{0.99} & 0.98 \\ \hline
CodeLlama-13B & \multicolumn{1}{c|}{0.92} & \multicolumn{1}{c|}{0.89} & \multicolumn{1}{c|}{0.94} & \multicolumn{1}{c|}{0.95} & \multicolumn{1}{c|}{0.95} & \multicolumn{1}{c|}{0.97} & \multicolumn{1}{c|}{0.97} & \multicolumn{1}{c|}{0.97} & \multicolumn{1}{c|}{0.96} & \multicolumn{1}{c|}{0.98} & 0.96 \\ \hline
CodeLlama-34B & \multicolumn{1}{c|}{\textbf{1.00}} & \multicolumn{1}{c|}{0.95} & \multicolumn{1}{c|}{0.98} & \multicolumn{1}{c|}{0.99} & \multicolumn{1}{c|}{0.98} & \multicolumn{1}{c|}{\textbf{1.00}} & \multicolumn{1}{c|}{\textbf{1.00}} & \multicolumn{1}{c|}{0.99} & \multicolumn{1}{c|}{\textbf{1.00}} & \multicolumn{1}{c|}{\textbf{1.00}} & \textbf{1.00} \\ \hline
Mistral-7B-v0.1 & \multicolumn{1}{c|}{0.05} & \multicolumn{1}{c|}{0.11} & \multicolumn{1}{c|}{0.10} & \multicolumn{1}{c|}{0.09} & \multicolumn{1}{c|}{0.13} & \multicolumn{1}{c|}{0.15} & \multicolumn{1}{c|}{0.12} & \multicolumn{1}{c|}{0.12} & \multicolumn{1}{c|}{0.11} & \multicolumn{1}{c|}{0.12} & 0.10 \\ \hline
Mistral-7B-v0.2 & \multicolumn{1}{c|}{0.02} & \multicolumn{1}{c|}{0.07} & \multicolumn{1}{c|}{0.10} & \multicolumn{1}{c|}{0.11} & \multicolumn{1}{c|}{0.09} & \multicolumn{1}{c|}{0.10} & \multicolumn{1}{c|}{0.12} & \multicolumn{1}{c|}{0.11} & \multicolumn{1}{c|}{0.12} & \multicolumn{1}{c|}{0.13} & 0.12 \\ \hline
StabilityAI-3B & \multicolumn{1}{c|}{0.00} & \multicolumn{1}{c|}{0.00} & \multicolumn{1}{c|}{0.00} & \multicolumn{1}{c|}{0.00} & \multicolumn{1}{c|}{0.00} & \multicolumn{1}{c|}{0.00} & \multicolumn{1}{c|}{0.00} & \multicolumn{1}{c|}{0.00} & \multicolumn{1}{c|}{0.00} & \multicolumn{1}{c|}{0.00} & 0.00 \\ \hline
\end{tabular}
    \end{subtable}%
    \vfill
    \begin{subtable}[t]{\textwidth}
        \caption{Optimal number of in-context samples for the function compositionality task (higher-order code from sequences of first-order code)}
        \label{subtab:2}    
\begin{tabular}{|c|ccccccccccc|}
\hline
\multirow{2}{*}{Model} & \multicolumn{11}{c|}{Number of In-Context Samples} \\ \cline{2-12} 
 & \multicolumn{1}{c|}{0} & \multicolumn{1}{c|}{1} & \multicolumn{1}{c|}{2} & \multicolumn{1}{c|}{3} & \multicolumn{1}{c|}{4} & \multicolumn{1}{c|}{5} & \multicolumn{1}{c|}{6} & \multicolumn{1}{c|}{7} & \multicolumn{1}{c|}{8} & \multicolumn{1}{c|}{9} & 10 \\ \hline
CodeLlama-7B & \multicolumn{1}{c|}{0.01} & \multicolumn{1}{c|}{0.80} & \multicolumn{1}{c|}{0.89} & \multicolumn{1}{c|}{0.92} & \multicolumn{1}{c|}{0.95} & \multicolumn{1}{c|}{0.94} & \multicolumn{1}{c|}{0.89} & \multicolumn{1}{c|}{0.93} & \multicolumn{1}{c|}{0.89} & \multicolumn{1}{c|}{0.92} & 0.85 \\ \hline
CodeLlama-13B & \multicolumn{1}{c|}{0.52} & \multicolumn{1}{c|}{0.73} & \multicolumn{1}{c|}{0.77} & \multicolumn{1}{c|}{0.80} & \multicolumn{1}{c|}{0.81} & \multicolumn{1}{c|}{0.82} & \multicolumn{1}{c|}{0.94} & \multicolumn{1}{c|}{0.93} & \multicolumn{1}{c|}{0.94} & \multicolumn{1}{c|}{0.93} & 0.91 \\ \hline
CodeLlama-34B & \multicolumn{1}{c|}{0.75} & \multicolumn{1}{c|}{0.67} & \multicolumn{1}{c|}{0.89} & \multicolumn{1}{c|}{0.92} & \multicolumn{1}{c|}{0.95} & \multicolumn{1}{c|}{0.99} & \multicolumn{1}{c|}{\textbf{1.00}} & \multicolumn{1}{c|}{\textbf{1.00}} & \multicolumn{1}{c|}{0.99} & \multicolumn{1}{c|}{\textbf{1.00}} & \textbf{1.00} \\ \hline
Mistral-7B-v0.1 & \multicolumn{1}{c|}{0.00} & \multicolumn{1}{c|}{0.27} & \multicolumn{1}{c|}{0.45} & \multicolumn{1}{c|}{0.48} & \multicolumn{1}{c|}{0.42} & \multicolumn{1}{c|}{0.34} & \multicolumn{1}{c|}{0.37} & \multicolumn{1}{c|}{0.34} & \multicolumn{1}{c|}{0.40} & \multicolumn{1}{c|}{0.35} & 0.32 \\ \hline
Mistral-7B-v0.2 & \multicolumn{1}{c|}{0.02} & \multicolumn{1}{c|}{0.24} & \multicolumn{1}{c|}{0.25} & \multicolumn{1}{c|}{0.23} & \multicolumn{1}{c|}{0.25} & \multicolumn{1}{c|}{0.20} & \multicolumn{1}{c|}{0.32} & \multicolumn{1}{c|}{0.35} & \multicolumn{1}{c|}{0.35} & \multicolumn{1}{c|}{0.29} & 0.30 \\ \hline
StabilityAI-3B & \multicolumn{1}{c|}{0.00} & \multicolumn{1}{c|}{0.00} & \multicolumn{1}{c|}{0.00} & \multicolumn{1}{c|}{0.00} & \multicolumn{1}{c|}{0.00} & \multicolumn{1}{c|}{0.00} & \multicolumn{1}{c|}{0.00} & \multicolumn{1}{c|}{0.00} & \multicolumn{1}{c|}{0.00} & \multicolumn{1}{c|}{0.00} & 0.00 \\ \hline
\end{tabular}
    \end{subtable}%
    \vfill
    \begin{subtable}[t]{\textwidth}
        \caption{Optimal number of in-context samples for the function compositionality task (higher-order optimal code)}
        \label{subtab:3}    
\begin{tabular}{|c|ccccccccccc|}
\hline
\multirow{2}{*}{Model} & \multicolumn{11}{c|}{Number of In-Context Samples} \\ \cline{2-12} 
 & \multicolumn{1}{c|}{0} & \multicolumn{1}{c|}{1} & \multicolumn{1}{c|}{2} & \multicolumn{1}{c|}{3} & \multicolumn{1}{c|}{4} & \multicolumn{1}{c|}{5} & \multicolumn{1}{c|}{6} & \multicolumn{1}{c|}{7} & \multicolumn{1}{c|}{8} & \multicolumn{1}{c|}{9} & 10 \\ \hline
CodeLlama-7B & \multicolumn{1}{c|}{0.01} & \multicolumn{1}{c|}{0.16} & \multicolumn{1}{c|}{0.22} & \multicolumn{1}{c|}{0.23} & \multicolumn{1}{c|}{0.29} & \multicolumn{1}{c|}{0.25} & \multicolumn{1}{c|}{0.27} & \multicolumn{1}{c|}{0.29} & \multicolumn{1}{c|}{0.829} & \multicolumn{1}{c|}{0.31} & 0.34 \\ \hline
CodeLlama-13B & \multicolumn{1}{c|}{0.52} & \multicolumn{1}{c|}{0.18} & \multicolumn{1}{c|}{0.28} & \multicolumn{1}{c|}{0.37} & \multicolumn{1}{c|}{0.42} & \multicolumn{1}{c|}{0.39} & \multicolumn{1}{c|}{0.46} & \multicolumn{1}{c|}{0.43} & \multicolumn{1}{c|}{0.45} & \multicolumn{1}{c|}{0.46} & 0.48 \\ \hline
CodeLlama-34B & \multicolumn{1}{c|}{\textbf{0.75}} & \multicolumn{1}{c|}{0.13} & \multicolumn{1}{c|}{0.30} & \multicolumn{1}{c|}{0.35} & \multicolumn{1}{c|}{0.43} & \multicolumn{1}{c|}{0.41} & \multicolumn{1}{c|}{0.45} & \multicolumn{1}{c|}{0.44} & \multicolumn{1}{c|}{0.42} & \multicolumn{1}{c|}{0.46} & 0.49 \\ \hline
Mistral-7B-v0.1 & \multicolumn{1}{c|}{0.00} & \multicolumn{1}{c|}{0.11} & \multicolumn{1}{c|}{0.10} & \multicolumn{1}{c|}{0.09} & \multicolumn{1}{c|}{0.13} & \multicolumn{1}{c|}{0.15} & \multicolumn{1}{c|}{0.12} & \multicolumn{1}{c|}{0.12} & \multicolumn{1}{c|}{0.11} & \multicolumn{1}{c|}{0.12} & 0.10 \\ \hline
Mistral-7B-v0.2 & \multicolumn{1}{c|}{0.02} & \multicolumn{1}{c|}{0.07} & \multicolumn{1}{c|}{0.10} & \multicolumn{1}{c|}{0.11} & \multicolumn{1}{c|}{0.09} & \multicolumn{1}{c|}{0.10} & \multicolumn{1}{c|}{0.12} & \multicolumn{1}{c|}{0.11} & \multicolumn{1}{c|}{0.12} & \multicolumn{1}{c|}{0.13} & 0.12 \\ \hline
StabilityAI-3B & \multicolumn{1}{c|}{0.00} & \multicolumn{1}{c|}{0.00} & \multicolumn{1}{c|}{0.00} & \multicolumn{1}{c|}{0.00} & \multicolumn{1}{c|}{0.00} & \multicolumn{1}{c|}{0.00} & \multicolumn{1}{c|}{0.00} & \multicolumn{1}{c|}{0.00} & \multicolumn{1}{c|}{0.00} & \multicolumn{1}{c|}{0.00} & 0.00 \\ \hline
\end{tabular}
    \end{subtable}
    \vfill
    
\begin{subtable}[t]{\textwidth}
\caption{Optimal number of in-context samples for the function repeatability task for the simple objects}
\label{tab:rb_so_ic_abl_study}
\begin{tabular}{|c|ccccccccccc|}
\hline
\multirow{2}{*}{Model} & \multicolumn{11}{c|}{Number of In-Context Samples} \\ \cline{2-12} 
 & \multicolumn{1}{c|}{0} & \multicolumn{1}{c|}{1} & \multicolumn{1}{c|}{2} & \multicolumn{1}{c|}{3} & \multicolumn{1}{c|}{4} & \multicolumn{1}{c|}{5} & \multicolumn{1}{c|}{6} & \multicolumn{1}{c|}{7} & \multicolumn{1}{c|}{8} & \multicolumn{1}{c|}{9} & 10 \\ \hline
CodeLlama-34B & \multicolumn{1}{c|}{0.01} & \multicolumn{1}{c|}{\textbf{0.89}} & \multicolumn{1}{c|}{0.63} & \multicolumn{1}{c|}{0.68} & \multicolumn{1}{c|}{0.70} & \multicolumn{1}{c|}{0.75} & \multicolumn{1}{c|}{0.82} & \multicolumn{1}{c|}{0.82} & \multicolumn{1}{c|}{0.82} & \multicolumn{1}{c|}{0.83} & 0.80 \\ \hline
\end{tabular}
\end{subtable}    
\vfill
\begin{subtable}[t]{\textwidth}
\caption{Optimal number of in-context samples for the function repeatability task (complex objects)}
\label{tab:rb_co_ic_abl_study}
\begin{tabular}{|c|ccccccccccc|}
\hline
\multirow{2}{*}{Model} & \multicolumn{11}{c|}{Number of In-Context Samples} \\ \cline{2-12} 
 & \multicolumn{1}{c|}{0} & \multicolumn{1}{c|}{1} & \multicolumn{1}{c|}{2} & \multicolumn{1}{c|}{3} & \multicolumn{1}{c|}{4} & \multicolumn{1}{c|}{5} & \multicolumn{1}{c|}{6} & \multicolumn{1}{c|}{7} & \multicolumn{1}{c|}{8} & \multicolumn{1}{c|}{9} & 10 \\ \hline
CodeLlama-34B & \multicolumn{1}{c|}{0.00} & \multicolumn{1}{c|}{0.01} & \multicolumn{1}{c|}{0.04} & \multicolumn{1}{c|}{0.02} & \multicolumn{1}{c|}{0.05} & \multicolumn{1}{c|}{0.05} & \multicolumn{1}{c|}{\textbf{0.07}} & \multicolumn{1}{c|}{\textbf{0.07}} & \multicolumn{1}{c|}{0.05} & \multicolumn{1}{c|}{0.05} & 0.06 \\ \hline
\end{tabular}
\vfill
\begin{subtable}[t]{\textwidth}
\centering
\caption{Impact of prompt structure elements on code generation LLM performance for the functional compositionality task. S: System Information, E: Environment Information, C: Context Information, T: Task Information, I: In-context Samples, I*: 5 In-context Samples, O: Other Information.}
\label{tab:prompt_structure}
\begin{tabular}{|c|c|}
\hline
Prompt Structure & CodeLlama-34b \\ \hline
S + E + C + T + O + I* & \textbf{0.41} \\ \hline
E + C + T + O + I* & 0.35 \\ \hline
S + C + T + O + I* & 0.39 \\ \hline
S + E + T + O + I* & 0.37 \\ \hline
S + E + C + O + I* & 0.33 \\ \hline
S + E + C + T + I* & 0.30 \\ \hline
\end{tabular}
    
\end{subtable}
    
\end{subtable}        

\end{table*}

\begin{table*}
\small
\centering
\caption{Overview of detailed error analysis for \textit{simple boards} across all the tasks}
\label{tab:error-info-details}
    \begin{subtable}[t]{\textwidth}
\caption{Detailed error analysis for the template-based instructions}
\label{tab:error-info-detail-tb}    
\begin{tabular}{|c|c|c|c|c|c|c|}
\hline
\begin{tabular}[c]{@{}c@{}}Board\\ Type\end{tabular} & \begin{tabular}[c]{@{}c@{}}Object\\ Type\end{tabular} & Task & Error Category & CodeLlama & GPT-4 & Claude-3 \\ \hline
\multirow{19}{*}{Simple} & \multirow{19}{*}{Simple} & \multirow{4}{*}{\begin{tabular}[c]{@{}c@{}}Property\\ Compositionality\end{tabular}}  & \textbf{Total} & \textbf{8} & \textbf{0} & \textbf{0} \\ \cline{4-7} 
 &  &  & Syntax Error & 3 & 0 & 0 \\ \cline{4-7} 
 &  &  & Mismatch Location & 3 & 0 & 0 \\ \cline{4-7} 
 &  &  & Mismatch Shape & 2 & 0 & 0 \\ \cline{3-7} 
 &  & \multirow{5}{*}{\begin{tabular}[c]{@{}c@{}}Function\\ Compositionality\\ (Using sequences of\\ first-order code)\end{tabular}} & \textbf{Total} & \textbf{5} & \textbf{0} & \textbf{9} \\ \cline{4-7} 
 &  &  & Syntax Error & 0 & 0 & 9 \\ \cline{4-7} 
 &  &  & Mismatch Count & 2 & 0 & 0 \\ \cline{4-7} 
 &  &  & Mismatch Location & 2 & 0 & 0 \\ \cline{4-7} 
 &  &  & Mismatch Shape & 1 & 0 & 0 \\ \cline{3-7} 
 &  & \multirow{10}{*}{\begin{tabular}[c]{@{}c@{}}Function\\ Compositionality\\ (Using higher-order\\ optimal code)\end{tabular}} & \textbf{Total} & \textbf{62} & \textbf{57} & \textbf{17} \\ \cline{4-7} 
 &  &  & Depth Mismatch & 16 & 34 & 11 \\ \cline{4-7} 
 &  &  & Dimensions Mismatch  & 3 & 5 & 4 \\ \cline{4-7} 
 &  &  & Bridge Placement & 6 & 3 & 0 \\ \cline{4-7} 
 &  &  & Value Error & 0 & 0 & 1 \\ \cline{4-7} 
 &  &  & Mismatch Location & 6 & 5 & 0 \\ \cline{4-7} 
 &  &  & Mismatch Count & 26 & 10 & 1 \\ \cline{4-7}
 &  &  & Mismatch Shape & 1 & 0 & 0 \\ \cline{4-7}
 &  &  & Same Shape Stacking & 2 & 0 & 0 \\ \cline{4-7} 
 &  &  & Same Shape At Alternate Levels & 2 & 0 & 0 \\ \hline
\end{tabular}

\end{subtable}%
    \vfill
    \begin{subtable}[t]{\textwidth}
\caption{Detailed error analysis for the human-authored instructions}
\label{tab:error-info-detail-hw}    
\begin{tabular}{|c|c|c|c|c|c|c|}
\hline
\begin{tabular}[c]{@{}c@{}}Board\\ Type\end{tabular} & \begin{tabular}[c]{@{}c@{}}Object\\ Type\end{tabular} & Task & Error Category & CodeLlama & GPT-4 & Claude-3 \\ \hline
\multirow{9}{*}{Simple} & \multirow{9}{*}{Simple} & \multirow{9}{*}{\begin{tabular}[c]{@{}c@{}}Function\\ Compositionality\\ (Using higher-order\\ optimal code)\end{tabular}} & \textbf{Total} & \textbf{94} & \textbf{108} & \textbf{74} \\ \cline{4-7} 
 &  &  & Depth Mismatch & 37 & 59 & 39 \\ \cline{4-7} 
 &  &  & Dimensions Mismatch & 3 & 9 & 5 \\ \cline{4-7} 
 &  &  & Bridge Placement & 15 & 4 & 5 \\ \cline{4-7} 
 &  &  & Value Error & 2 & 5 & 4 \\ \cline{4-7} 
 &  &  & Syntax Error & 0 & 0 & 4 \\ \cline{4-7}  
 &  &  & Mismatch Location & 9 & 6 & 1 \\ \cline{4-7} 
 &  &  & Mismatch Color & 1 & 0 & 0 \\ \cline{4-7} 
 &  &  & Mismatch Count & 23 & 24 & 16 \\ \cline{4-7}
 &  &  & Same Shape Stacking & 2 & 0 & 0 \\ \cline{4-7} 
 &  &  & Same Shape At Alternate Levels & 2 & 1 & 0 \\ \hline
\end{tabular}

\end{subtable}%

    \vfill
    \begin{subtable}[t]{\textwidth}
\caption{Detailed error analysis for the model-generated instructions}
\label{tab:error-info-detail-mg}    
\begin{tabular}{|c|c|c|c|c|c|c|}
\hline
\begin{tabular}[c]{@{}c@{}}Board\\ Type\end{tabular} & \begin{tabular}[c]{@{}c@{}}Object\\ Type\end{tabular} & Task & Error Category & CodeLlama & GPT-4 & Claude-3 \\ \hline
\multirow{9}{*}{Simple} & \multirow{9}{*}{Simple} & \multirow{9}{*}{\begin{tabular}[c]{@{}c@{}}Function\\ Compositionality\\ (Using higher-order\\ optimal code)\end{tabular}} & \textbf{Total} & \textbf{125} & \textbf{101} & \textbf{100} \\ \cline{4-7} 
 &  &  & Depth Mismatch & 43 & 88 & 82 \\ \cline{4-7} 
 &  &  & Dimensions Mismatch & 5 & 1 & 0 \\ \cline{4-7} 
 &  &  & Bridge Placement & 14 & 0 & 1 \\ \cline{4-7} 
 &  &  & Value Error & 11 & 3 & 4 \\ \cline{4-7}
 &  &  & Name Error & 1 & 0 & 0 \\ \cline{4-7}  
 &  &  & Syntax Error & 8 & 0 & 7 \\ \cline{4-7}  
 &  &  & Not On Top Of Screw & 11 & 0 & 0 \\ \cline{4-7} 
 &  &  & Mismatch Location & 13 & 0 & 0 \\ \cline{4-7} 
 &  &  & Mismatch Count & 8 & 9 & 4 \\ \cline{4-7}
 &  &  & Mismatch Shape & 3 & 0 & 0 \\ \cline{4-7} 
 &  &  & Same Shape Stacking & 4 & 0 & 2 \\ \cline{4-7} 
 &  &  & Same Shape At Alternate Levels & 4 & 0 & 0 \\ \hline
\end{tabular}

\end{subtable}%
\end{table*}
\begin{table*}
\footnotesize
\centering
\caption{Overview of detailed error analysis for \textit{regular boards} across all the tasks for the tmplate-based and human-authored instructions}
\label{tab:error-info-details-rb}
    \begin{subtable}[t]{\textwidth}
\footnotesize
\caption{Detailed error analysis for the template-based instructions}
\label{tab:error-info-detail-rb-tb}
\begin{tabular}{|c|c|c|c|c|c|c|}
\hline
\begin{tabular}[c]{@{}c@{}}Board\\ Type\end{tabular} & \begin{tabular}[c]{@{}c@{}}Object\\ Type\end{tabular} & Task & Error Category & CodeLlama-34b & GPT-4 & Claude-3 \\ \hline
\multirow{21}{*}{Regular} & \multirow{8}{*}{Simple} & \multirow{21}{*}{\begin{tabular}[c]{@{}c@{}}Function\\ Repeatability\end{tabular}} & \textbf{Total} & \textbf{33} & \textbf{0} & \textbf{18} \\ \cline{4-7} 
 &  &  & Depth Mismatch & 4 & 0 & 1 \\ \cline{4-7} 
 &  &  & Bridge Placement & 5 & 0 & 0 \\ \cline{4-7} 
 &  &  & Mismatch Location & 18 & 0 & 0 \\ \cline{4-7} 
 &  &  & Syntax Error & 0 & 0 & 16 \\ \cline{4-7} 
 &  &  & Name Error & 2 & 0 & 0 \\ \cline{4-7} 
 &  &  & Same Shape At Alternate Levels & 3 & 0 & 1 \\ \cline{4-7} 
 &  &  & Same Shape Stacking & 1 & 0 & 0 \\ \cline{2-2} \cline{4-7} 
 & \multirow{13}{*}{Complex} &  & \textbf{Total} & \textbf{119} & \textbf{91} & \textbf{117} \\ \cline{4-7} 
 &  &  & Depth Mismatch & 4 & 1 & 2 \\ \cline{4-7} 
 &  &  & Dimensions Mismatch & 3 & 0 & 7 \\ \cline{4-7} 
 &  &  & Bridge Placement & 7 & 4 & 2 \\ \cline{4-7} 
 &  &  & Value Error & 11 & 10 & 5 \\ \cline{4-7} 
 &  &  & Syntax Error & 2 & 0 & 92 \\ \cline{4-7} 
 &  &  & Name Error & 2 & 0 & 0 \\ \cline{4-7} 
 &  &  & Mismatch Location & 72 & 43 & 6 \\ \cline{4-7} 
 &  &  & Mismatch Count & 9 & 25 & 2 \\ \cline{4-7} 
 &  &  & Mismatch Color & 0 & 1 & 0 \\ \cline{4-7}  
 &  &  & Mismatch Shape & 1 & 5 & 1  \\ \cline{4-7} 
 &  &  & Not On Top Of Screw & 2 & 1 & 0 \\ \cline{4-7} 
 &  &  & Same Shape At Alternate Levels & 6 & 1 & 0 \\ \hline
\end{tabular}
\end{subtable}%
    \vfill
    \begin{subtable}[t]{\textwidth}
\caption{Detailed error analysis for the human-authored instructions}
\label{tab:error-info-detail-rb-hw}    
\begin{tabular}{|c|c|c|c|c|c|c|}
\hline
\begin{tabular}[c]{@{}c@{}}Board\\ Type\end{tabular} & \begin{tabular}[c]{@{}c@{}}Object\\ Type\end{tabular} & Task & Error Category & CodeLlama-34b & GPT-4 & Claude-3 \\ \hline
\multirow{25}{*}{Regular} & \multirow{12}{*}{Simple} & \multirow{25}{*}{\begin{tabular}[c]{@{}c@{}}Function\\ Repeatability\end{tabular}} & \textbf{Total} & \textbf{121} & \textbf{90} & \textbf{98} \\ \cline{4-7} 
 &  &  & Depth Mismatch & 5 & 0 & 2 \\ \cline{4-7} 
 &  &  & Dimensions Mismatch & 7 & 5 & 0 \\ \cline{4-7} 
 &  &  & Bridge Placement & 6 & 2 & 2 \\ \cline{4-7} 
 &  &  & Value Error & 8 & 3 & 2 \\ \cline{4-7} 
 &  &  & Syntax Error & 6 & 0 & 72 \\ \cline{4-7} 
 &  &  & Name Error & 2 & 0 & 1 \\ \cline{4-7} 
 &  &  & Key Error & 12 & 0 & 1 \\ \cline{4-7} 
 &  &  & Type Error & 4 & 0 & 0 \\ \cline{4-7} 
 &  &  & Mismatch Location & 32 & 18 & 2 \\ \cline{4-7}
 &  &  & Mismatch Count & 14 & 6 & 2 \\ \cline{4-7} 
 &  &  & Mismatch Color & 24 & 55 & 14 \\ \cline{4-7} 
 &  &  & Same Shape At Alternate Levels & 1 & 1 & 0 \\  \cline{2-2} \cline{4-7}
 & \multirow{12}{*}{Complex} &  & \textbf{Total} & \textbf{121} & \textbf{94} & \textbf{125} \\ \cline{4-7} 
 &  &  & Depth Mismatch & 1 & 0 & 2 \\ \cline{4-7} 
 &  &  & Dimensions Mismatch & 13 & 6 & 3 \\ \cline{4-7} 
 &  &  & Bridge Placement & 3 & 0 & 3 \\ \cline{4-7} 
 &  &  & Value Error & 23 & 9 & 2 \\ \cline{4-7} 
 &  &  & Syntax Error & 6 & 6 & 103 \\ \cline{4-7} 
 &  &  & Name Error & 7 & 0 & 0 \\ \cline{4-7} 
 &  &  & Key Error & 22 & 0 & 1 \\ \cline{4-7} 
 &  &  & Mismatch Location & 19 & 17 & 6 \\ \cline{4-7} 
 &  &  & Mismatch Color & 20 & 53 & 4 \\ \cline{4-7}
 &  &  & Mismatch Count & 7 & 3 & 0 \\ \cline{4-7}
 &  &  & Same Shape Stacking & 0 & 0 & 1 \\ \hline
\end{tabular}
\end{subtable}%
\end{table*}   
\begin{table*}
\footnotesize
\centering
\caption{Overview of detailed error analysis for \textit{regular boards} across all the tasks for the model-generated instructions}
\label{tab:error-info-detail-rb-mg}
    \begin{subtable}[t]{\textwidth}
%\caption{Detailed error analysis for the model-generated instructions}
%\label{tab:error-info-detail-rb-mg}    
\begin{tabular}{|c|c|c|c|c|c|c|}
\hline
\begin{tabular}[c]{@{}c@{}}Board\\ Type\end{tabular} & \begin{tabular}[c]{@{}c@{}}Object\\ Type\end{tabular} & Task & Error Category & CodeLlama-34b & GPT-4 & Claude-3 \\ \hline
\multirow{12}{*}{Regular} & \multirow{12}{*}{Simple} & \multirow{12}{*}{\begin{tabular}[c]{@{}c@{}}Function\\ Repeatability\end{tabular}} & \textbf{Total} & \textbf{130} & \textbf{113} & \textbf{87} \\ \cline{4-7} 
 &  &  & Depth Mismatch & 2 & 0 & 0 \\ \cline{4-7} 
 &  &  & Dimensions Mismatch & 41 & 13 & 1 \\ \cline{4-7} 
 &  &  & Bridge Placement & 0 & 2 & 0 \\ \cline{4-7} 
 &  &  & Mismatch Location & 66 & 85 & 33 \\ \cline{4-7} 
 &  &  & Mismatch Count & 0 & 2 & 0 \\ \cline{4-7}
 &  &  & Mismatch Shape & 0 & 1 & 0 \\ \cline{4-7}
 &  &  & Name Error & 2 & 0 & 0 \\ \cline{4-7}  
 &  &  & Value Error & 11 & 2 & 0 \\ \cline{4-7}  
 &  &  & Syntax Error & 5 & 8 & 51 \\ \cline{4-7}  
 &  &  & Not On Top Of Screw & 1 & 0 & 1 \\ \cline{4-7}   
&  &  & Same Shape At Alternate Levels & 2 & 0 & 1 \\  \hline 
 \end{tabular}
\end{subtable}%
\end{table*}    

\clearpage
\newpage

\begin{figure}[htb]
\small
\centering
    \begin{minipage}[t]{\textwidth}
    \begin{subfigure}[t]{\textwidth}
    \small
    \centering    
        \begin{mdframed}
            \centering 
            \begin{lstlisting}
{%- if instruction_index == 0 -%}
These are the step-by-step instructions to build {{ data.combo_name }}.{{ " " }}
{%- endif -%}
{%- if orientation -%}
place a {{ color }} {{ shape }} {{ orientation }}ly in the {{ x }} row, {{ y }} column
{%- else -%}
place a {{ color }} {{ shape }} in the {{ x }} row, {{ y }} column
{%- endif -%}
            \end{lstlisting}
        \end{mdframed}
        \caption{Multi-turn Instruction template}
        \label{fig:sb_so_multiturn}
    \end{subfigure}
    
    \begin{subfigure}[t]{\textwidth}
    \small
    \centering    
        \begin{mdframed}
            \centering 
            \begin{lstlisting}
These are the instructions to build {{ data.combo_name }}.{{ " " }}
{%- for i in range(data.shapes|length) -%}
Place a {{ data.colors[i] }} {{ data.shapes[i] }} {%- if data.orientations[i] %} {{ data.orientations[i] }}ly {%- endif %} in the {{ x[i] }} row, {{ y[i] }} column.{% if not loop.last %} {% endif %}
{%- endfor -%}                
            \end{lstlisting}
        \end{mdframed}
        \caption{Single-turn Instruction template}
        \label{fig:sb_so_singleturn}
    \end{subfigure}  
    \end{minipage}
    \caption{\textit{Jinja2} templates for generating multi-turn and single-turn instructions for simple boards. \textbf{Top}: Defines the template grammar used for atomic placement of a component in each turn. \textbf{Bottom}: Defines the template grammar used for multiple component arrangement in a sequence.}
    \label{fig:instruction_templates_sb}    
\end{figure}

\begin{figure}[htb]
\small
\centering
    \begin{minipage}[t]{\textwidth}
    \begin{subfigure}[t]{\textwidth}
    \small
    \centering    
        \begin{mdframed}
            \centering 
            \begin{lstlisting}
Place a '{{combo_name}}' object in the first, fourth, and seventh columns of the first row. Then, repeat this pattern of placement in the remaining rows. Use only these colors: {{ colors }} for the '{{combo_name}}' object.
            \end{lstlisting}
        \end{mdframed}
        \caption{arrangement-1}
        \label{fig:rb_so_1}
    \end{subfigure}
    
    \begin{subfigure}[t]{\textwidth}
    \small
    \centering    
        \begin{mdframed}
            \centering 
            \begin{lstlisting}
Place a '{{combo_name}}' object in the first, fourth, and seventh rows of the first column. Then, repeat this placement pattern in the remaining columns.  Use only these colors: {{ colors }} for the '{{combo_name}}' object.         
            \end{lstlisting}
        \end{mdframed}
        \caption{arrangement-2}
        \label{fig:rb_so_2}
    \end{subfigure}   

    \begin{subfigure}[t]{\textwidth}
    \small
    \centering    
        \begin{mdframed}
            \centering 
            \begin{lstlisting}
Starting from the top-left corner, fill the grid with '{{ combo_name }}' objects diagonally.  Use only these colors: {{ colors }} for the '{{combo_name}}' object.         
            \end{lstlisting}
        \end{mdframed}
        \caption{arrangement-3}
        \label{fig:rb_so_3}
    \end{subfigure} 

    \begin{subfigure}[t]{\textwidth}
    \small
    \centering    
        \begin{mdframed}
            \centering 
            \begin{lstlisting}
Place a '{{ combo_name }}' object at all the corners of the grid.  Use only these colors: {{ colors }} for the '{{combo_name}}' object.          
            \end{lstlisting}
        \end{mdframed}
        \caption{arrangement-4}
        \label{fig:rb_so_4}
    \end{subfigure} 

    \begin{subfigure}[t]{\textwidth}
    \small
    \centering
        \begin{mdframed}
            \centering 
            \begin{lstlisting}
Place a '{{combo_name}}' object in the first, and fifth columns of the first row. Then, repeat this placement pattern in the fifth row.  Use only these colors: {{ colors }} for the '{{combo_name}}' object.         
            \end{lstlisting}
        \end{mdframed}
        \caption{arrangement-5}
        \label{fig:rb_so_5}
    \end{subfigure}     
    \end{minipage}
    \caption{\textit{Jinja2} templates for the construction of simple objects in regular boards. Each template grammar refers to a particular sequence arrangement.}
    \label{fig:instruction_templates_rb_so}    
\end{figure}

\clearpage
\newpage

\begin{figure}[htb]
\small
\centering

    \begin{subfigure}[b]{\textwidth}
    \small
    \centering
        \begin{mdframed}
            \begin{lstlisting}
Start from the top-left corner and diagonally place '{{ combo_name }}' objects, each taking a {{ occupied_rows }}x{{ occupied_columns }} space. Continue until there's insufficient space at the grid's end for a full object.  Use only these colors: {{ colors }} for the '{{combo_name}}' object.
            \end{lstlisting}
        \end{mdframed}
        \caption{arrangement-1}
        \label{fig:sub1-11}
    \end{subfigure}
    
    \begin{subfigure}[b]{\textwidth}
    \small
    \centering
        \begin{mdframed}
            \begin{lstlisting}
Starting from the top-left corner, place the '{{ combo_name }}' object in alternating columns of the first row, each occupying a {{ occupied_rows }}x{{ occupied_columns }} space.  Continue until there's insufficient space at the grid's end for a full object. Then, repeat this pattern in alternating rows.  Use only these colors: {{ colors }} for the '{{combo_name}}' object.        
            \end{lstlisting}
        \end{mdframed}
        \caption{arrangement-2}
        \label{fig:sub2-2}
    \end{subfigure}   

    \begin{subfigure}[b]{\textwidth}
    \small
    \centering
        \begin{mdframed}
            \begin{lstlisting}
Fill the first column with the '{{ combo_name }}' object, each occupying a {{ occupied_rows }}x{{ occupied_columns }} space.  Continue until there's insufficient space at the grid's end for a full object. Then repeat this action in every alternate column.  Use only these colors: {{ colors }} for the '{{combo_name}}' object.         
            \end{lstlisting}
        \end{mdframed}
        \caption{arrangement-3}
        \label{fig:sub2-22}
    \end{subfigure} 

    \begin{subfigure}[b]{\textwidth}
    \small
    \centering
        \begin{mdframed}
            \begin{lstlisting}
Fill the fourth column with the '{{ combo_name }}' object, each occupying a {{ occupied_rows }}x{{ occupied_columns }} space. Continue until there's insufficient space at the grid's end for a full object.  Use only these colors: {{ colors }} for the '{{combo_name}}' object.         
            \end{lstlisting}
        \end{mdframed}
        \caption{arrangement-4}
        \label{fig:sub2-222}
    \end{subfigure} 
  
    \caption{\textit{Jinja2} templates for the construction of complex objects in regular boards.}
    \label{fig:instruction_templates_rb_co}    
\end{figure}

\begin{figure}
\small
\centering
    \begin{subfigure}[b]{\textwidth}
    \small
    \centering
        \begin{mdframed}
            \begin{lstlisting}
def {{ combo_name }}(board, colors, x, y):
    shapes = ['{{ shapes[0] }}', '{{ shapes[1] }}']
    for shape, color in zip(shapes, colors):
            put(board, shape, color, x, y)            
            \end{lstlisting}
        \end{mdframed}
        \caption{Stack two shapes}
        \label{fig:sub2-2-2}
    \end{subfigure}

    \begin{subfigure}[b]{\textwidth}
    \small
    \centering
        \begin{mdframed}
            \begin{lstlisting}
def {{ combo_name }}(board, colors, x, y):
    shapes = ['{{ shapes[0] }}', '{{ shapes[1] }}', '{{ shapes[2] }}']
    for shape, color in zip(shapes, colors):
            put(board, shape, color, x, y)             
            \end{lstlisting}
        \end{mdframed}
        \caption{Stack three shapes}
        \label{fig:sub2-2-22}
    \end{subfigure}   

    \begin{subfigure}[b]{\textwidth}
    \small
    \centering
        \begin{mdframed}
            \begin{lstlisting}
def {{ combo_name }}(board, colors, x, y):
    shapes = ['{{ shapes[0] }}', '{{ shapes[1] }}', 'bridge-h']
    for shape, color, dx, dy in zip(shapes, colors, [0, 0, 0], [0, 1, 0]):
            put(board, shape, color, x + dx, y + dy)           
            \end{lstlisting}
        \end{mdframed}
        \caption{Place two shapes adjacently on a row and stack a horizontal bridge on top of them}
        \label{fig:sub3-33}
    \end{subfigure}     

    \begin{subfigure}[b]{\textwidth}
    \small
    \centering
        \begin{mdframed}
            \begin{lstlisting}
def {{ combo_name }}(board, colors, x, y):
    shapes = ['{{ shapes[0] }}', '{{ shapes[1] }}', 'bridge-v']
    for shape, color, dx, dy in zip(shapes, colors, [0, 1, 0], [0, 0, 0]):
            put(board, shape, color, x + dx, y + dy)         
            \end{lstlisting}
        \end{mdframed}
        \caption{Place two shapes vertically in a column and stack a vertical bridge on top of them}
        \label{fig:sub4-4-444}
    \end{subfigure}      
     
    \caption{\textit{Jinja2} templates for the code generation for the component arrangements}
    \label{fig:code_templates_1}    
\end{figure}

\clearpage
\newpage
\begin{figure}[htb]
\small
\centering
    \begin{subfigure}[b]{\textwidth}
    \small
    \centering
        \begin{mdframed}
            \begin{lstlisting}
def {{ combo_name }}(board, colors, x, y):
    shapes = ['bridge-h', '{{ shapes[1] }}', '{{ shapes[2] }}']
    for shape, color, dx, dy in zip(shapes, colors, [0, 0, 0], [0, 1, 1]):
            put(board, shape, color, x + dx, y + dy)       
            \end{lstlisting}
        \end{mdframed}
        \caption{Place a horizontal bridge and stack two shapes on its right side}
        \label{fig:sub555}
    \end{subfigure}  

    \begin{subfigure}[b]{\textwidth}
    \small
    \centering
        \begin{mdframed}
            \begin{lstlisting}
def {{ combo_name }}(board, colors, x, y):
    shapes = ['bridge-v', '{{ shapes[1] }}', '{{ shapes[2] }}']
    for shape, color, dx, dy in zip(shapes, colors, [0, 1, 1], [0, 0, 0]):
            put(board, shape, color, x + dx, y + dy)      
            \end{lstlisting}
        \end{mdframed}
        \caption{Place a vertical bridge and stack two shapes on the bottom side of it}
        \label{fig:sub666}
    \end{subfigure}    

    \begin{subfigure}[b]{\textwidth}
    \small
    \centering
        \begin{mdframed}
            \begin{lstlisting}
def {{ combo_name }}(board, colors, x, y):
    shapes = ['{{ shapes[0] }}', '{{ shapes[1] }}', '{{ shapes[2] }}', '{{ shapes[3] }}']
    for shape, color in zip(shapes, colors):
            put(board, shape, color, x, y)
            \end{lstlisting}
        \end{mdframed}
        \caption{Stack four shapes}
        \label{fig:sub11}
    \end{subfigure}
    
    \begin{subfigure}[b]{\textwidth}
    \small
    \centering
        \begin{mdframed}
            \begin{lstlisting}
def {{ combo_name }}(board, colors, x, y):
    shapes = ['{{ shapes[0] }}', '{{ shapes[1] }}', 'bridge-h', '{{ shapes[3] }}']
    for shape, color, dx, dy in zip(shapes, colors, [0, 0, 0, 0], [0, 1, 0, 0]):
            put(board, shape, color, x + dx, y + dy)            
            \end{lstlisting}
        \end{mdframed}
        \caption{Place two shapes side by side on a row, then stack a horizontal bridge and an additional shape on top of these two shapes}
        \label{fig:sub2-2-222}
    \end{subfigure}   

    \caption{\textit{Jinja2} templates for the code generation for the component arrangements}
    \label{fig:code_templates_2}    
\end{figure}

\begin{figure}[htb]
\small
\centering
    \begin{subfigure}[b]{\textwidth}
    \small
    \centering
        \begin{mdframed}
            \begin{lstlisting}
def {{ combo_name }}(board, colors, x, y):
    shapes = ['{{ shapes[0] }}', '{{ shapes[1] }}', 'bridge-h', '{{ shapes[3] }}']
    for shape, color, dx, dy in zip(shapes, colors, [0, 0, 0, 0], [0, 1, 0, 1]):
            put(board, shape, color, x + dx, y + dy)         
            \end{lstlisting}
        \end{mdframed}
        \caption{Place two shapes adjacently on a row and stack a horizontal bridge along with a fourth shape on top of these two shapes}
        \label{fig:sub33}
    \end{subfigure}  

    \begin{subfigure}[b]{\textwidth}
    \small
    \centering
        \begin{mdframed}
            \begin{lstlisting}
def {{ combo_name }}(board, colors, x, y):
    shapes = ['{{ shapes[0] }}', '{{ shapes[1] }}', 'bridge-v', '{{ shapes[3] }}']
    for shape, color, dx, dy in zip(shapes, colors, [0, 1, 0, 0], [0, 0, 0, 0]):
            put(board, shape, color, x + dx, y + dy)      
            \end{lstlisting}
        \end{mdframed}
        \caption{Place two shapes vertically in a column and stack a vertical bridge and an additional shape on top of these two shapes}
        \label{fig:sub44}
    \end{subfigure}

    \begin{subfigure}[b]{\textwidth}
    \small
    \centering
        \begin{mdframed}
            \begin{lstlisting}
def {{ combo_name }}(board, colors, x, y):
    shapes = ['{{ shapes[0] }}', '{{ shapes[1] }}', 'bridge-v', '{{ shapes[3] }}']
    for shape, color, dx, dy in zip(shapes, colors, [0, 1, 0, 1], [0, 0, 0, 0]):
            put(board, shape, color, x + dx, y + dy)     
            \end{lstlisting}
        \end{mdframed}
        \caption{Place two shapes vertically in a column and stack a vertical bridge along with a fourth shape on top of these two shapes}
        \label{fig:sub5-5}
    \end{subfigure}      

    \caption{\textit{Jinja2} templates for the code generation for the component arrangements}
    \label{fig:code_templates_3}    
\end{figure}

\clearpage
\newpage

\begin{figure}[htb]
\small
\centering
    \begin{subfigure}[b]{\textwidth}
    \small
    \centering
        \begin{mdframed}
            \begin{lstlisting}
def {{ combo_name }}(board, colors, x, y):
    shapes = ['bridge-h', '{{ shapes[1] }}', 'bridge-v', '{{ shapes[3] }}']
    for shape, color, dx, dy in zip(shapes, colors, [0, 1, 0, 0], [0, 1, 1, 1]):
            put(board, shape, color, x + dx, y + dy) 
            \end{lstlisting}
        \end{mdframed}
        \caption{Placement and stacking of a combination that includes a horizontal bridge, a vertical bridge, and two other shapes}
        \label{fig:sub66}
    \end{subfigure}    

    \begin{subfigure}[b]{\textwidth}
    \small
    \centering
        \begin{mdframed}
            \begin{lstlisting}
def {{ combo_name }}(board, colors, x, y):
    shapes = ['bridge-v', '{{ shapes[1] }}', 'bridge-h', '{{ shapes[3] }}']
    for shape, color, dx, dy in zip(shapes, colors, [0, 1, 1, 1], [0, 1, 0, 0]):
            put(board, shape, color, x + dx, y + dy)
            \end{lstlisting}
        \end{mdframed}
        \caption{Placement and stacking of a combination that includes a vertical bridge, a horizontal bridge, and two other shapes}
        \label{fig:sub7}
    \end{subfigure}   

    \begin{subfigure}[b]{\textwidth}
    \small
    \centering
        \begin{mdframed}
            \begin{lstlisting}
def {{ combo_name }}(board, colors, x, y):
    shapes = ['bridge-h', 'bridge-h', 'bridge-v', 'bridge-v']
    for shape, color, dx, dy in zip(shapes, colors, [0, 1, 0, 0], [0, 0, 0, 1]):
            put(board, shape, color, x + dx, y + dy)

            \end{lstlisting}
        \end{mdframed}
        \caption{Placement and stacking of combinations of horizontal and vertical bridges}
        \label{fig:sub8}
    \end{subfigure}     
    
    \begin{subfigure}[b]{\textwidth}
    \small
    \centering
        \begin{mdframed}
            \begin{lstlisting}
def {{ combo_name }}(board, colors, x, y):
    shapes = ['bridge-v', 'bridge-v', 'bridge-h', 'bridge-h']
    for shape, color, dx, dy in zip(shapes, colors, [0, 0, 0, 1], [0, 1, 0, 0]):
            put(board, shape, color, x + dx, y + dy)
            \end{lstlisting}
        \end{mdframed}
        \caption{Placement and stacking of combinations of vertical and horizontal bridges}
        \label{fig:sub111}
    \end{subfigure}

    \caption{\textit{Jinja2} templates for the code generation for the component arrangements}
    \label{fig:code_templates_4}    
\end{figure}

\begin{figure}[htb]
\small
\centering

    \begin{subfigure}[b]{\textwidth}
    \small
    \centering
        \begin{mdframed}
            \begin{lstlisting}
def {{ combo_name }}(board, colors, x, y):
    shapes = ['bridge-h', '{{ shapes[1] }}', 'bridge-v', '{{ shapes[3] }}', '{{ shapes[4] }}']
    for shape, color, dx, dy in zip(shapes, colors, [0, 1, 0, 0, 0], [0, 1, 1, 1, 1]):
            put(board, shape, color, x + dx, y + dy)            
            \end{lstlisting}
        \end{mdframed}
        \caption{Placement and Stacking of horizontal bridge, vertical bridge and a combination of two other shapes}
        \label{fig:sub22}
    \end{subfigure}    

    \begin{subfigure}[b]{\textwidth}
    \small
    \centering
        \begin{mdframed}
            \begin{lstlisting}
def {{ combo_name }}(board, colors, x, y):
    shapes = ['{{ shapes[0] }}', '{{ shapes[1] }}', 'bridge-h', '{{ shapes[3] }}', '{{ shapes[4] }}']
    for shape, color, dx, dy in zip(shapes, colors, [0, 0, 0, 0, 0], [0, 1, 0, 0, 0]):
            put(board, shape, color, x + dx, y + dy)       
            \end{lstlisting}
        \end{mdframed}
        \caption{Place two shapes adjacently on a row and stack a horizontal bridge along with two other shapes on top of these two shapes}
        \label{fig:sub333}
    \end{subfigure}   

    \begin{subfigure}[b]{\textwidth}
    \small
    \centering
        \begin{mdframed}
            \begin{lstlisting}
def {{ combo_name }}(board, colors, x, y):
    shapes = ['{{ shapes[0] }}', '{{ shapes[1] }}', 'bridge-v', '{{ shapes[3] }}', '{{ shapes[4] }}']
    for shape, color, dx, dy in zip(shapes, colors, [0, 1, 0, 0, 0], [0, 0, 0, 0, 0]):
            put(board, shape, color, x + dx, y + dy)        
            \end{lstlisting}
        \end{mdframed}
        \caption{Place two shapes vertically in a row and stack a vertical bridge along with two additional shapes on top of these two shapes}
        \label{fig:sub444}
    \end{subfigure}        
    
    \begin{subfigure}[b]{\textwidth}
    \small
    \centering
        \begin{mdframed}
            \begin{lstlisting}
for row in range({{ num_rows }}):
    for col in [0, 3, 6]:
        {{ combo_name }}(board, colors={{ colors }},x=row, y=col)
            \end{lstlisting}
        \end{mdframed}
        \caption{Repeat a simple object in the first, fourth, and seventh columns across all the rows in the grid}
        \label{fig:sub1-1}
    \end{subfigure}

    \caption{\textit{Jinja2} templates for the code generation for the component arrangements}
    \label{fig:code_templates_5}    
\end{figure}

\clearpage
\newpage

\begin{figure}[htb]
\small
\centering  

    \begin{minipage}[t]{\textwidth}
    
    \begin{subfigure}[t]{\textwidth}
    \small
    \centering
        \begin{mdframed}
            \begin{lstlisting}
for row in [0, 3, 6]:
    for col in range({{ num_cols }}):
        {{ combo_name }}(board, colors={{ colors }},x=row, y=col)           
            \end{lstlisting}
        \end{mdframed}
        \caption{Repeat a simple object in the first, fourth, and seventh rows across all columns of the grid}
        \label{fig:sub222}
    \end{subfigure}   

    \begin{subfigure}[t]{\textwidth}
    \small
    \centering
        \begin{mdframed}
            \begin{lstlisting}
for row in range({{ num_rows }}):
    for col in range({{ num_cols }}):
        if row == col:
            {{ combo_name }}(board, colors={{ colors }},x=row, y=col)     
            \end{lstlisting}
        \end{mdframed}
        \caption{Fill the grid by placing a simple object along the diagonal}
        \label{fig:sub3-3}
    \end{subfigure}   

    \begin{subfigure}[t]{\textwidth}
    \small
    \centering
        \begin{mdframed}
            \begin{lstlisting}
for row, col in [[0,0], [0,{{ num_cols-1 }}], [{{ num_rows-1 }}, 0], [{{ num_rows-1 }}, {{ num_cols-1 }}]]:
    {{ combo_name }}(board, colors={{ colors }}, x=row, y=col)       
            \end{lstlisting}
        \end{mdframed}
        \caption{Place a simple object at all the corners of the grid}
        \label{fig:sub4-4}
    \end{subfigure}  

    \begin{subfigure}[t]{\textwidth}
    \small
    \centering
        \begin{mdframed}
            \begin{lstlisting}
for row in [0, 4]:
    for col in [0, 4]:
        {{ combo_name }}(board, colors={{ colors }}, x=row, y=col)      
            \end{lstlisting}
        \end{mdframed}
        \caption{Fill the entire fourth column and the entire fourth row of the grid with a simple object}
        \label{fig:sub4-44}
    \end{subfigure}  
    \end{minipage}

    \caption{\textit{Jinja2} templates for the code generation for the component arrangements}
    \label{fig:code_templates_6}    
\end{figure}    
\vspace{-10mm}

\begin{figure}[htb]
\small
\centering  

    \begin{minipage}[t]{\textwidth}
    \begin{subfigure}[t]{\textwidth}
    \small    
    \centering
        \begin{mdframed}
            \begin{lstlisting}
for row in range(0, {{ num_rows }}, {{ 2+min_rows-1 }}):
    for col in range(0, {{ num_cols }}, {{ 2+min_cols-1 }}):
        {{ combo_name }}(board, colors={{ colors }},x=row, y=col)
      
            \end{lstlisting}
        \end{mdframed}
        \caption{Place a complex object in alternating columns of the first row, and replicate this pattern in alternating rows throughout the grid.}
        \label{fig:sub4-444}
    \end{subfigure}      

    \begin{subfigure}[t]{\textwidth}
    \small
    \centering
        \begin{mdframed}
            \begin{lstlisting}
for row in range(0, {{ num_rows }}, {{ min_rows }}):
    for col in range(0, {{ num_cols }}, {{ 2+min_cols-1 }}):
        {{ combo_name }}(board, colors={{ colors }},x=row, y=col)
            \end{lstlisting}
        \end{mdframed}
        \caption{Place a complex object in alternating rows of the first column. Apply this pattern to alternating columns across the grid}
        \label{fig:sub4-4-4}
    \end{subfigure}     

    \begin{subfigure}[t]{\textwidth}
    \small
    \centering
        \begin{mdframed}
            \begin{lstlisting}
for row in range(0, {{ num_rows }}, {{ min_rows }}):
    {{ combo_name }}(board, colors={{ colors }}, x=row, y=3)
            \end{lstlisting}
        \end{mdframed}
        \caption{Repeat a complex object in the fourth column of the grid}
        \label{fig:sub4-4-44}
    \end{subfigure}     
    \end{minipage}

    \caption{\textit{Jinja2} templates for the code generation for the component arrangements}
    \label{fig:code_templates_7}    
\end{figure}    

\end{document}